\documentclass[sigconf,pbalance]{acmart_modified}

\usepackage{bbm}
\usepackage{array}
\usepackage{amsmath}
\usepackage{graphicx}
\usepackage{tabularx}
\usepackage{booktabs}
\usepackage{multirow}
\usepackage{subcaption}
\usepackage{caption}
\usepackage{dblfloatfix}
\usepackage[most]{tcolorbox}
\usepackage{fvextra}
\usepackage{paralist}
\usepackage[table]{xcolor}
\usepackage{enumitem}
\usepackage{makecell}
\usepackage{circuitikz}

\definecolor{no_reasoning}{rgb}{0.0, 0.4, 0.8}
\definecolor{reasoning}{rgb}{0.0, 0.6, 0.4}

\definecolor{cellgreen}{RGB}{158, 210, 150}   
\definecolor{cellred}{RGB}{235, 150, 140}     
\definecolor{darkorange}{RGB}{200,100,0}

\newboolean{anonymous}
\setboolean{anonymous}{false} 

\newcommand{\squeezeSection}{}
\newcommand{\squeezeSubSection}{}

\newtcolorbox{prompt}[1][]{
    enhanced,
    drop shadow={black!50!white},
    coltitle=black,
    top=4pt,
    bottom=-1pt,
    left=2pt,
    right=2pt,
    attach boxed title to top left={xshift=1.5em,yshift=-\tcboxedtitleheight/2},
    boxed title style={size=small, colback=lightgray},
    fonttitle=\bfseries\scriptsize,
    title={#1}
}

\newcommand{\llmspm}{LLM-SPICE\-Mixer}
\newcommand{\spm}{SPICE\-Mixer}

\copyrightyear{2026}
\acmYear{2026}
\setcopyright{cc}
\setcctype{by}
\acmConference[MLCAD '26]{2026 ACM/IEEE International Symposium on Machine Learning for CAD}{September 07--09, 2026}{Jeju Island, Republic of Korea}
\acmBooktitle{2026 ACM/IEEE International Symposium on Machine Learning for CAD (MLCAD '26), September 07--09, 2026, Jeju Island, Republic of Korea}
\acmDOI{10.1145/3831599.3840324}
\acmISBN{979-8-4007-2878-5/2026/09}

\begin{document}

\title{Spicing up Genetic Netlist Generation with LLMs}

\ifthenelse{\boolean{anonymous}}{
  \author{Anonymous Authors}
  \affiliation{%
    \institution{Anonymous Institution(s)}
    \country{}}
  \renewcommand{\shortauthors}{Anonymous}
  \setcopyright{acmlicensed}
  \copyrightyear{2026}
  \acmYear{2026}
  \acmDOI{XXXXXXX.XXXXXXX}
  \acmConference[MLCAD '26]{International Symposium on Machine Learning for CAD 2026}{September 7-9, 2026}{Jeju, South Korea}
}{
    \author{Stefan Uhlich\textsuperscript{1}, Yağız Gençer\textsuperscript{1,2}, Andrea Bonetti\textsuperscript{1},\linebreak Arun Venkitaraman\textsuperscript{1}, Chia-Yu Hsieh\textsuperscript{1}, Eisaku Ohbuchi\textsuperscript{3}, Lorenzo Servadei\textsuperscript{1,4}}
    \affiliation{%
    \institution{\textsuperscript{1}\textit{Sony AI, Switzerland} \quad
    \textsuperscript{2}\textit{EPFL, Switzerland} \quad \textsuperscript{3}\textit{Sony Semiconductor Solutions, Japan} \quad
    \textsuperscript{4}\textit{TU Munich, Germany}}\country{}}
    \renewcommand{\shortauthors}{Uhlich, Gençer, Bonetti, Venkitaraman, Hsieh, Ohbuchi, Servadei}
}

\begin{abstract}
Analog circuit topology synthesis remains challenging because useful designs occupy a tiny fraction of a combinatorial search space, and small structural changes can induce highly nonlinear changes in behavior. Evolutionary algorithms are attractive because they can optimize over discrete circuit topologies using only black-box evaluations, but they often require many SPICE simulations and may converge prematurely. We introduce \llmspm{}, a hybrid synthesis framework that augments genetic netlist generation with IGEL (\underline{I}nspiration-\underline{G}uided \underline{E}volution with \underline{L}LMs), an LLM-based proposal operator. During search, IGEL prompts an LLM with high-performing circuits from the elite set and instructs it to generate a new SPICE netlist, which is then evaluated by SPICE and selected using the same reward mechanism as conventional genetic operators. Thus, the LLM contributes structured topology proposals while simulation remains the source of truth. We evaluate \llmspm{} on a challenging benchmark task: synthesizing transistor-level circuits that implement a discriminant function for Iris classification. Compared with the genetic framework without LLM guidance, \llmspm{} improves the median final training reward by $8.4\%$ and the median validation-selected test reward by $8.8\%$. The best validation-selected circuit achieves $93.3\%$ test accuracy at the nominal \texttt{tt} corner and $85.9\%$ average test accuracy across 17 process, voltage, and temperature corners.
\squeezeSection
\end{abstract}



\maketitle

\squeezeSection\section{Introduction}

\begin{figure*}
    \resizebox{\linewidth}{!}{\includegraphics[trim=20 0 0 0]{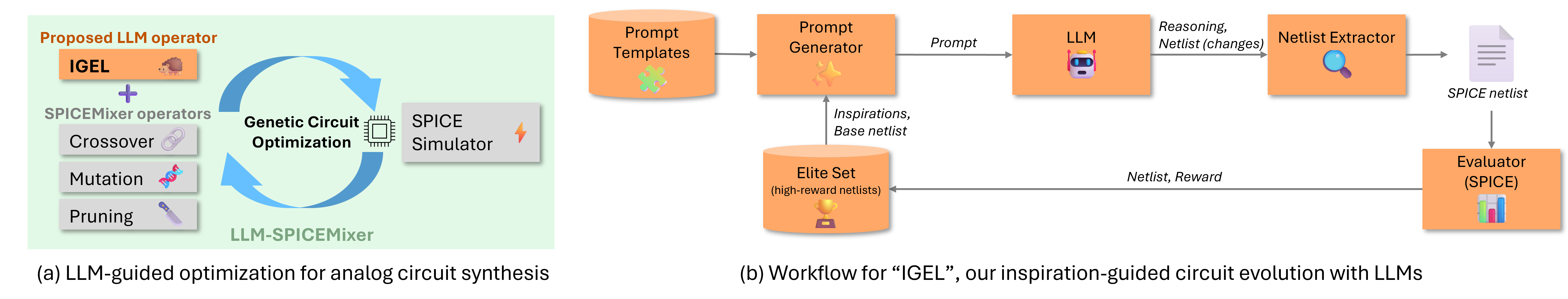}}
    \vspace{-0.7cm}
    \caption{Overview of \llmspm{}: (a) LLM-augmented genetic circuit optimization loop. (b) IGEL prompts the LLM with inspiration circuits, then evaluates the proposed netlist with SPICE; it is added to the elite set if good enough.}
    \label{fig:OverallSystem}
    \vspace{-0.3cm}
\end{figure*}

Analog circuit design remains challenging because small structural modifications can lead to highly nonlinear and non-smooth changes in behavior~\cite{razavi2017analogcmos}. Circuit synthesis requires both topology selection, which determines devices and interconnections, and sizing, which sets parameters such as transistor widths and lengths. For fixed topologies, substantial progress has been made using Bayesian optimization~\cite{lyu2017efficient,gu2024tss}, reinforcement learning~\cite{settaluri2020autockt,wang2020gcn,budak2021dnn,kim2025ppaas}, and large language models (LLMs)~\cite{somayaji2025llm,liu2025eesizer,ahmadzadeh2025anaflow}. Topology synthesis remains harder because it involves discrete structural choices and is strongly coupled with sizing. Consequently, many successful approaches focus on well-studied circuit families such as operational amplifiers~\cite{lu2022topology,zhao2022analog,chen2023total,poddar2024data,shen2025into}, filters~\cite{koza1996automated,hu2005open,rojec2018analog}, power converters~\cite{fan2021specification,vijayaraghavan2025autocircuit,gao2026powergenie}, or other common circuit classes~\cite{gao2025analoggenie,lai2025analogcoder,lai2026analogcoder,kim2026analogtobi}. This leaves open how to synthesize useful circuits for less standardized tasks, where no canonical topology or family-specific search space is available.

Evolutionary algorithms are attractive for open-ended topology synthesis because they optimize candidate circuits directly through SPICE simulation. However, purely genetic search can require many evaluations and may converge prematurely. LLMs offer a complementary capability: they can generate structured text and may capture useful regularities from circuit descriptions and netlists. Yet asking an LLM to design a complete circuit from scratch is unreliable, especially for non-standard tasks where memorized templates are unlikely to apply. This suggests a hybrid strategy: use the LLM not as a standalone circuit designer, but as a proposal mechanism inside a simulation-driven evolutionary loop.

Motivated by this idea, we introduce \llmspm{}, a circuit synthesis method that extends \spm{}~\cite{uhlich2025spicemixer} with IGEL, an LLM-based proposal operator for \emph{\underline{I}nspiration-\underline{G}uided Evolution with \underline{L}LMs}. During search, IGEL samples high-quality circuits from the elite set and provides them to the LLM as inspirations. The LLM then proposes a new candidate netlist, which is evaluated by SPICE and selected using the same reward mechanism as conventional genetic operators such as mutation, crossover, and pruning. Thus, the LLM injects structured topology variations, while SPICE simulation provides the performance evaluation. The goal is not to replace genetic search, but to make it less prone to premature convergence by adding informed and diverse proposals.

We evaluate \llmspm{} on an analog discriminant-function synthesis task based on Iris classification. The goal is to synthesize a transistor-level circuit whose input voltages encode the four Iris features and whose output voltages represent the three class scores. This task is useful as a benchmark because it is specified by data-dependent classification behavior rather than standard analog design targets, and suitable topologies are not known a priori. Thus, the method must discover compact transistor-level structures rather than tune a known circuit template.

\noindent In summary, the contributions of this paper are as follows:
\begin{itemize}[noitemsep, topsep=0pt, leftmargin=*]
    \item We introduce \llmspm{}, a hybrid synthesis method that augments genetic netlist generation with IGEL, an LLM-based proposal operator that generates candidate circuits from elite netlists within a SPICE-driven evolutionary loop.
    \item We propose an analog discriminant-function synthesis benchmark based on the Iris dataset and use it to evaluate open-ended transistor-level topology synthesis.
    \item We present an empirical study of prompting strategies, decoding settings, model choices, and operator mixtures, showing that IGEL improves search performance when used as a complement to conventional genetic operators.
\end{itemize}

\noindent The paper is organized as follows. Sec.~\ref{sec:related_work} reviews related work. Sec.~\ref{sec:llm_spicemixer} introduces \llmspm{} and IGEL. Sec.~\ref{sec:synthesis_task} describes the analog discriminant-function synthesis task and evaluation setup. Sec.~\ref{sec:results} presents results, including baseline comparisons and ablations. Finally, Sec.~\ref{sec:conclusions_and_outlook} concludes and outlines future work.

\squeezeSection\section{Related Work}
\label{sec:related_work}
Analog design automation has seen strong progress for sizing fixed topologies using Bayesian optimization, reinforcement learning, and, more recently, LLM-based optimization~\cite{lyu2017efficient,gu2024tss,settaluri2020autockt,wang2020gcn,budak2021dnn,kim2025ppaas,somayaji2025llm,liu2025eesizer,ahmadzadeh2025anaflow}. Topology synthesis is more challenging: early work relied on evolutionary search and genetic programming for joint topology-and-sizing optimization~\cite{koza1996automated,hu2005open,rojec2018analog}, while recent methods often focus on specific circuit families, particularly operational amplifiers and power converters~\cite{lu2022topology,zhao2022analog,chen2023total,poddar2024data,shen2025into,fan2021specification,gao2026powergenie}. Other approaches address more general circuit generation using graph-based, generative, or LLM-based methods~\cite{uhlich2024graco,gao2025analoggenie,kim2026analogtobi,vijayaraghavan2025autocircuit,lai2025analogcoder,lai2026analogcoder}. Agentic LLM frameworks such as AnalogAgent~\cite{bao2026analogagent} have also introduced multi-agent workflows for iterative analog circuit generation. These methods typically use learned models or LLMs as primary generators or design agents, whereas our approach uses the LLM only as one proposal operator inside a genetic search.

A separate line of work integrates LLMs into evolutionary search. FunSearch and related systems use an LLM to propose candidates that are evaluated and selected within an iterative search loop~\cite{romera2024funsearch,liu2023ael,liu2024eoh,ye2024reevo,vanstein2024llamea,novikov2025alphaevolve,zhang2026trajectory}. Our method follows this paradigm, but transfers it to transistor-level analog synthesis: it builds on \spm{}~\cite{uhlich2025spicemixer}, which evolves SPICE netlists directly, and extends it with IGEL, an LLM-based proposal operator that generates candidates from elite netlists. In contrast to general program-search systems, our candidates are physical circuits whose quality must be determined by SPICE simulation under task-specific testbenches and process corners. Related uses of LLM-generated SPICE netlists also appear in SpiceFuzz~\cite{ren2026spicefuzz}, but there the goal is simulator fuzzing rather than reward-driven functional synthesis. To the best of our knowledge, \mbox{\llmspm{}} is the first method to embed LLM-generated netlist proposals into an evolutionary loop for analog circuit synthesis.

\squeezeSection\section{\llmspm{}}
\label{sec:llm_spicemixer}
We first briefly review \spm{}~\cite{uhlich2025spicemixer}. We then introduce IGEL, our LLM-based proposal operator, which enriches the elite set during genetic search. The overall system is illustrated in Fig.~\ref{fig:OverallSystem}.

\squeezeSubSection\subsection{Recap of \spm{}}
\label{subsec:spicemixer:sec:llm_spicemixer}
\spm{} is a circuit synthesis method based on a genetic algorithm that evolves SPICE netlists directly. Rather than relying on an abstract graph representation or a hand-crafted chromosome encoding, it treats the netlist itself as the genome. This makes the method naturally compatible with arbitrary components and subcircuits, and therefore easy to adapt to different circuit libraries and process design kits.

\spm{} repeatedly applies one of three proposal operators to generate new candidate circuits. The first operator is \emph{crossover}, which combines two elite netlists line by line. The second operator is \emph{mutation}, which combines one elite netlist with a randomly sampled netlist to introduce new structural variations while preserving useful parts of a strong parent. The third operator is \emph{pruning}, which merges compatible component definitions so that the resulting circuit can become smaller and more compact.

Each newly generated netlist is then evaluated by SPICE simulation under a task-specific testbench, and its performance is mapped to a scalar reward that serves as the fitness value. \spm{} maintains an elite set containing the best circuits found so far, and parent circuits are sampled from this set using a rank-based roulette-wheel strategy. This biases the search toward high-quality solutions while still preserving diversity.

\squeezeSubSection\subsection{Inspiration-Guided Evolution with LLMs}
\label{subsec:llm_proposal_operator:sec:llm_spicemixer}
Because circuits can be represented as netlists, LLMs are a natural choice for circuit synthesis. However, their effectiveness depends strongly on how they are used. One option is to ask the LLM to generate a complete circuit directly, as in AnalogCoder~\cite{lai2025analogcoder} and AnalogCoder-Pro~\cite{lai2026analogcoder}. In our setting, however, this strategy is often ineffective. Our goal is to discover novel circuit topologies from uncommon circuit families, and in such cases the model often reproduces the same canonical solution~\cite{wright2025epistemic,yun2025price}, even across multiple interaction rounds and even when given feedback on circuit performance as we will show in Sec.~\ref{subsec:igel_only:sec:results}. Although some methods can mitigate this behavior~\cite{zhang2025verbalized}, it remains a general limitation. Another possibility would be to use the LLM as a verification tool, either as a pre-check before SPICE simulation or as a surrogate model for estimating circuit quality. We expect this to be unreliable as well, because the model would need to assess novel circuits for novel tasks solely from their netlists.

We therefore use the LLM as a proposal operator within a genetic search process, which we call IGEL. In this sense, the LLM is used in a manner conceptually similar to coding-agent systems such as AlphaEvolve~\cite{novikov2025alphaevolve}. Since LLMs are strong at code generation, we expect them to be well suited for proposing new candidate netlists, especially when diverse prompt formulations are used. To reduce generation collapse, we do not rely on IGEL alone. Instead, we combine it with the original \spm{} operators---crossover, mutation, and pruning---as shown in Fig.~\ref{fig:OverallSystem}a. In this way, the different operators can complement one another, since each of them can further refine a solution that was produced by another operator in an earlier step.

\subsubsection*{General Approach}
Fig.~\ref{fig:OverallSystem}b illustrates our use of the LLM. From the current elite set, which contains the best solutions found so far, we sample $K=3$ netlists using rank-based roulette-wheel sampling. We populate a prompt template with these netlists and their reward scores, and pass the resulting prompt to the LLM. The model then analyzes the given circuits and, when requested by the prompt, first produces a short plan describing how they should be modified. It subsequently outputs a new candidate netlist. We extract this netlist from the LLM response and evaluate it in a SPICE testbench.

\subsubsection*{Netlist Preprocessing}
We apply two preprocessing steps to improve the effectiveness of the method. First, we identify and remove circuits that share the same topology and differ only in parameter values. By ensuring that the LLM only sees structurally distinct inspirations, we encourage it to propose topological modifications rather than merely adjust transistor sizes. In preliminary experiments, we observed that without this step, the genetic algorithm often converged prematurely to an elite set containing only a single topology with minor sizing variations, many of which were produced by IGEL. Once this occurred, the search rarely recovered from the resulting collapse and often failed to discover better circuits.

Second, rather than providing the original netlists directly, we convert PDK-specific transistor instances, in our case from the SkyWater PDK~\cite{skywater130pdk}, into a simplified MOS representation. In this process, we remove the explicit bulk connection and rename nets to more descriptive names, for example \texttt{net\_supply\_0} to \texttt{net\_vdd} and \texttt{0} to \texttt{net\_gnd}. For example, the line

\vspace{0.1cm}
\noindent
\colorbox{gray!15}{%
  \parbox{0.99\linewidth}{%
    \small\texttt{%
      X6 net\_internal\_1 net\_internal\_0 net\_supply\_0 net\_supply\_0\\
      + sky130\_fd\_pr\_\_pfet\_01v8 w=20 l=1%
    }%
  }%
}

\vspace{0.2cm}
\noindent is rewritten as
\vspace{0.2cm}

\noindent
\colorbox{gray!15}{%
  \parbox{0.99\linewidth}{%
    \small\texttt{%
      M6 net\_internal\_1 net\_internal\_0 net\_vdd PMOS w=20 l=1%
    }%
  }%
}
\vspace{-0.2cm}

\noindent This preprocessing reduces potential confusion caused by uncommon component and net names that may have been underrepresented during LLM training. It also removes ambiguities, for example when \texttt{0} could refer either to ground or to a parameter value.

\subsubsection*{Prompt Templates}
To analyze the effect of prompting on synthesis performance, we compare prompt templates that vary along two dimensions:
\begin{itemize}[noitemsep, topsep=0pt, leftmargin=*]
    \item \emph{``raw'' vs. ``diff''}. In the ``raw'' setting, the LLM generates a complete candidate netlist. In the ``diff'' setting, the LLM specifies only the lines that should be changed relative to a given base netlist.
    \item \emph{``with reasoning'' vs. ``without reasoning''}. In the ``with reasoning'' setting, the prompt asks the LLM to first analyze the given inspirations and describe a modification strategy before generating the output. In the ``without reasoning'' setting, the LLM is instructed to produce the output directly, without an explicit planning step.
\end{itemize}
These two design choices result in four prompt templates. In Appx.~\ref{appx:prompt_and_output}, Fig.~\ref{fig:full_style_template} and Fig.~\ref{fig:diff_style_template} show example prompts together with the corresponding LLM outputs.

We expect both explicit reasoning and edit-based generation to improve performance. Asking the LLM to analyze the inspiration circuits before proposing a new netlist may help it identify useful structural patterns. Likewise, the ``diff'' format is well suited to genetic search because it encourages smaller, more local modifications of high-quality circuits. Based on this intuition, prompt templates with reasoning and structured edits appear to be strong candidates, although the best design remains an empirical question. As shown in Sec.~\ref{subsec:prompt:sec:results}, explicit reasoning is particularly beneficial, whereas the difference between ``raw'' and ``diff'' is comparatively small. Overall, the best performance is obtained by an alternating scheme that cycles between the reasoning-based ``raw'' and ``diff'' formats, suggesting that the additional diversity introduced by using two prompt styles is advantageous.

\squeezeSection\section{Analog Discriminant-Function Synthesis}
\label{sec:synthesis_task}
We evaluate \llmspm{} on the synthesis of an analog circuit that implements a discriminant function for the Iris classification task~\cite{fisher1936use}. This is an interesting non-standard synthesis problem, because the objective is not to reproduce a known circuit class, but to realize a classifier directly as an analog circuit. Hence, IGEL cannot rely on memorized solutions from the LLM, but must instead infer useful patterns from the inspirations.

\squeezeSubSection\subsection{Task Setup}
\label{subsec:analog_discriminant}
To implement the discriminant function for the Iris dataset, the circuit has four input nets corresponding to the four Iris features (\emph{sepal length (cm)}, \emph{sepal width (cm)}, \emph{petal length (cm)}, and \emph{petal width (cm)}), and three output nets corresponding to the three classes (\emph{setosa}, \emph{versicolor}, and \emph{virginica}). All four features are normalized by min-max scaling computed on the training set. During circuit evaluation, the normalized feature values are mapped to input voltages in the range $[0\,V, 1.8\,V]$. For a given input sample, the predicted class is determined by the output node with the highest voltage.

We use the standard Iris dataset, which contains 150 samples. The data are split into training, validation, and test sets using a stratified 60/20/20 split, yielding 90 training samples, 30 validation samples, and 30 test samples. This split is used such that the search optimizes the training reward, the validation set is used to select the best circuits found during the search, and the test set is reserved exclusively for final evaluation. To evaluate a candidate circuit efficiently, we present all samples from the training and validation splits within a single transient simulation. Each feature is applied through a piecewise-linear voltage source. Each sample is presented for $8\,$ns, and the circuit outputs are read at the end of this time window. To reduce artifacts caused by a fixed sample order, we shuffle each split three times with different random seeds and concatenate the resulting sequences. This produces longer input streams while preserving the class distribution. The same shuffling-and-concatenation procedure is applied to the held-out test split during final evaluation.

The synthesized circuits are built entirely from SkyWater transistors~\cite{skywater130pdk}. In addition to the topology, that is, the transistor interconnection pattern, the search space also includes transistor width and length as continuous parameters for each device. We attach a small output load capacitance of $10\,$fF to each output node to model the input capacitance of a subsequent stage.

\squeezeSubSection\subsection{Reward and Multi-Corner Evaluation}
\label{subsec:reward_multicorner}
Each candidate netlist is evaluated with Ngspice~\cite{ngspice46} under multiple process, voltage, and temperature conditions. We consider one nominal corner, \texttt{tt} at $1.8\,$V and $25^\circ$C, and 16 extreme corners formed by all combinations of process corners \texttt{ff}, \texttt{ss}, \texttt{sf}, and \texttt{fs}, supply voltages $1.62\,$V and $1.98\,$V, and temperatures $0^\circ$C and $85^\circ$C. In total, each circuit is evaluated on 17 corners, and the results are averaged. This encourages the search to find solutions that are robust across operating conditions and reduces the risk that a circuit performs well only because of artifacts of the idealized SPICE compact models.

For each sample, the ideal output is a one-hot voltage vector: the correct class should be close to $1.8\,$V, while the two incorrect classes should be close to $0\,$V. Let $\mathbf{y}_i \in \mathbb{R}^3$ denote the simulated output voltages for sample $i$, and let $\mathbf{t}_i \in \{0\,\text{V},1.8\,\text{V}\}^3$ denote the corresponding one-hot target. For one data split and one corner, we define the reward as $R = \frac{1}{2}(A + M)$, where $A$ is the classification accuracy,
\begin{equation}
A = \frac{1}{N} \sum_{i=1}^{N} \mathbbm{1}\!\left[\arg\max_j y_{i,j} = \arg\max_j t_{i,j}\right],
\label{eq:acc}
\end{equation}
and $M$ is the target-voltage score,
\begin{equation}
M = 1 - \frac{1}{3N} \sum_{i=1}^{N} \left\lVert \mathbf{y}_i - \mathbf{t}_i \right\rVert_1.
\label{eq:margin}
\end{equation}
Here, $N$ is the number of samples in the split, including the repeated shuffled sequences described above, and $\lVert .\rVert_1$ denotes the 1-norm. The maximum possible reward is therefore $R = 1$, while in practice $R$ is smaller. The first term, $A$ in \eqref{eq:acc}, rewards correct predictions. The second term, $M$ in \eqref{eq:margin}, rewards large voltage separation by encouraging the correct output to move toward the supply voltage and the incorrect outputs toward ground. This is important because two circuits can achieve the same accuracy while exhibiting very different output confidence. To obtain a robust classifier, we therefore seek a large output separation score $M$.

Furthermore, we apply two additional reward penalties to guide the search toward compact circuits that are not only functional but also physically plausible:
\begin{itemize}[noitemsep, topsep=0pt, leftmargin=*]
\item \emph{Validity penalty.} A penalty of $0.05$ is applied for each violated structural validity check and may therefore be incurred multiple times by the same netlist. These checks require that all three output nets are present, transistor bulk terminals are connected correctly, input nets are connected only to transistor gates, no floating nets exist, output nets are connected only to transistor drain or source terminals, and supply or ground nets are not connected to transistor gates.
\item \emph{Size penalty.} A penalty of $0.0025 \times N_\text{transistors}$ is subtracted to favor smaller circuits, where $N_\text{transistors}$ denotes the number of transistors in the netlist, that is, the number of netlist lines.
\end{itemize}
All reward values reported in this paper are penalized rewards, including both the validity and size penalties described above.

For every candidate, we compute training and validation rewards for each corner and then average them over all 17 corners. The genetic search uses the averaged training reward as the fitness value, while the averaged validation metric is recorded and later used to select the circuit with the highest validation reward. This follows standard machine learning practice and helps reduce overfitting to the training split. If a simulation fails or the output traces cannot be parsed correctly, the candidate receives a reward of $-1$, which discourages invalid circuits.
\squeezeSection\section{Results}
\label{sec:results}
In the experiments below, \emph{IGEL} denotes the concrete instantiation of our LLM-based proposal operator within \llmspm{}. At each IGEL step, we sample $K=3$ high-quality circuits from the current elite set using rank-based roulette-wheel sampling, preprocess them into a simplified MOS representation, and provide them to the LLM as \emph{inspirations}. The model then proposes one new candidate netlist, which is converted back to the SkyWater PDK representation, evaluated by SPICE, and inserted into the evolutionary search. Unless stated otherwise in the ablation studies, we use the following IGEL configuration: reasoning-enabled prompts with an alternating ``raw''/``diff'' format (cf., Sec.~\ref{subsec:prompt:sec:results}), balanced decoding (cf., Sec.~\ref{subsec:decoding:sec:results}), and Qwen3.5 27B as the proposal model (cf., Sec.~\ref{subsec:model:sec:results}). Each synthesis run is executed for 131{,}072 proposal steps, where each step generates and evaluates one candidate circuit. The four proposal operators are sampled with relative weights $1:1:1:0.5$ for crossover, mutation, pruning, and IGEL, respectively. Thus, IGEL is invoked only half as often as each conventional operator, reducing the computational cost of LLM inference.

For all experiments, we use vLLM~\cite{kwon2023efficient} to serve the LLMs on GPU servers. Models with up to 12B parameters (Gemma3 270M, 1B, 4B, and 12B, as well as Qwen3.5 9B) are run on Ada6000 GPUs, whereas larger models (Gemma3 27B and Qwen3.5 27B) are run on H200 GPUs. To reduce statistical noise, each method or configuration is evaluated over nine independent runs.

We first compare \llmspm{} with the baseline that does not use IGEL, namely \spm{}, and show that IGEL substantially improves performance. We then compare the best synthesized circuits with standard machine learning baselines, specifically logistic regression and a single-hidden-layer MLP. Finally, we perform ablation studies to isolate the effect of the main design choices in \llmspm{}, including the prompt template, decoding settings, and model family and size.

\squeezeSubSection\subsection{Comparison with Evolutionary Baselines}
\label{subsec:comparison:sec:results}

\begin{figure}
    \resizebox{\linewidth}{!}{\includegraphics[trim=0 20 0 0]{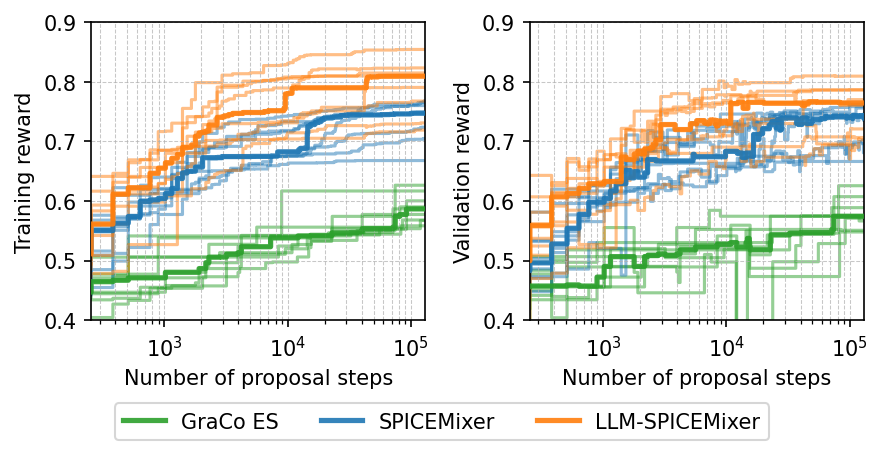}}
    \caption{Training and validation rewards of the best-so-far circuit in each run. Circuits are selected by training reward and then evaluated on validation. Each run uses $131{,}072$ proposal steps; thicker lines indicate medians over nine runs.}
    \label{fig:evolution_curves}
    \vspace{-0.3cm}
\end{figure}

We first compare \llmspm{} against two evolutionary baselines. The first is GraCo ES~\cite{uhlich2024graco}, which uses a graph neural network (GNN) to sequentially construct a graph representation of the synthesized circuit, while an evolutionary strategy (ES)~\cite{salimans2017evolution} is used to optimize the GNN parameters. The second baseline is \spm{}~\cite{uhlich2025spicemixer}, which was reviewed in Sec.~\ref{subsec:spicemixer:sec:llm_spicemixer}.

The evolution of the reward on the training and validation splits is shown in Fig.~\ref{fig:evolution_curves}, and the final training rewards are summarized in Tab.~\ref{tab:method_comparison}. GraCo ES performs worst: the search quickly stalls, and we observe that it produces circuits with highly similar topologies, leading to generation collapse and a median reward of only $0.587$. \spm{} performs substantially better, achieving a median reward of $0.747$. \llmspm{} yields the strongest overall performance. It improves the median reward to $0.810$ ($+0.063$ over \spm{}) and also discovers the best overall circuit, with a reward of $0.855$ ($+0.087$ over the best circuit found by \spm{}).

To assess whether this improvement is statistically significant, we perform a one-sided unpaired permutation test on the results of the nine independent runs. The test shows that the median improvement of $0.063$ is statistically significant, with a $p$-value of $0.0052$. These results indicate that adding the LLM-based proposal operator substantially improves the search and leads to better final solutions.

Fig.~\ref{fig:evolution_curves} also shows the evolution of the reward curves on the training and validation splits. Each plot reports the reward of the best circuit found so far according to the training set and evaluates that same circuit on the validation set. We observe that \llmspm{} consistently outperforms \spm{}, demonstrating that IGEL benefits the synthesis process. We also observe a train--validation gap: the final median reward reaches $0.81$ on the training set, compared to approximately $0.77$ on the validation set. Nevertheless, the overall ranking remains unchanged, and \llmspm{} also produces the best circuits on the validation split.

Using the validation reward, we select the best circuit from each run and evaluate it on the test split. Tab.~\ref{tab:test_reward_best_val_pick} summarizes these results. Again, \llmspm{} performs best, with a median test reward of $0.807$, which is $+0.065$ higher than \spm{}.

\begin{table}
    \caption{Final best training reward for different methods. Values are computed over nine independent runs.}
    \label{tab:method_comparison}
    \vspace{-0.35cm}
    \centering
    \resizebox{\linewidth}{!}{%
    \small
    \renewcommand{\arraystretch}{1.15}
    \setlength{\tabcolsep}{4pt}
    \begin{tabular}{@{}rw{c}{2.0cm}w{c}{2.0cm}w{c}{2.0cm}}
    \toprule
     & \textbf{GraCo ES} \cite{uhlich2024graco} & \textbf{\spm{}} \cite{uhlich2025spicemixer} & \textbf{\llmspm{}} \\
    \midrule
    \textbf{Average} $\pm$ \textbf{Std. Dev.}
      & \cellcolor{cellred!82}  0.588 $\pm$ 0.022
      & \cellcolor{cellgreen!7} 0.737 $\pm$ 0.031
      & \cellcolor{cellgreen!52} 0.799 $\pm$ 0.039 \\
    \textbf{Minimum}
      & \cellcolor{cellred!100} 0.558
      & \cellcolor{cellred!35}  0.668
      & \cellcolor{cellred!5}   0.719 \\
    \textbf{Median}
      & \cellcolor{cellred!83}   0.587
      & \cellcolor{cellgreen!15} 0.747
      & \cellcolor{cellgreen!65} 0.810 \\
    \textbf{Maximum}
      & \cellcolor{cellred!60}    0.626
      & \cellcolor{cellgreen!31}  0.768
      & \cellcolor{cellgreen!100} 0.855 \\
    \bottomrule
    \end{tabular}}
\end{table}

\begin{table}
    \caption{Test rewards obtained by selecting the checkpoints with the highest validation reward. Values are computed over nine independent runs.}
    \label{tab:test_reward_best_val_pick}
    \vspace{-0.35cm}
    \centering
    \resizebox{\linewidth}{!}{%
    \small
    \renewcommand{\arraystretch}{1.15}
    \setlength{\tabcolsep}{4pt}
    \begin{tabular}{@{}rw{c}{2.0cm}w{c}{2.0cm}w{c}{2.0cm}}
    \toprule
     & \textbf{GraCo ES} \cite{uhlich2024graco} & \textbf{\spm{}} \cite{uhlich2025spicemixer} & \textbf{\llmspm{}} \\
    \midrule
    \textbf{Average} $\pm$ \textbf{Std. Dev.}
      & \cellcolor{cellred!68}   0.627 $\pm$ 0.025
      & \cellcolor{cellgreen!9}  0.730 $\pm$ 0.031
      & \cellcolor{cellgreen!59} 0.782 $\pm$ 0.041 \\
    \textbf{Minimum}
      & \cellcolor{cellred!100} 0.580
      & \cellcolor{cellred!41}  0.667
      & \cellcolor{cellred!5}   0.715 \\
    \textbf{Median}
      & \cellcolor{cellred!71}   0.622
      & \cellcolor{cellgreen!20} 0.742
      & \cellcolor{cellgreen!82} 0.807 \\
    \textbf{Maximum}
      & \cellcolor{cellred!39}   0.670
      & \cellcolor{cellgreen!50} 0.773
      & \cellcolor{cellgreen!100} 0.824 \\
    \bottomrule
    \end{tabular}}
\end{table}

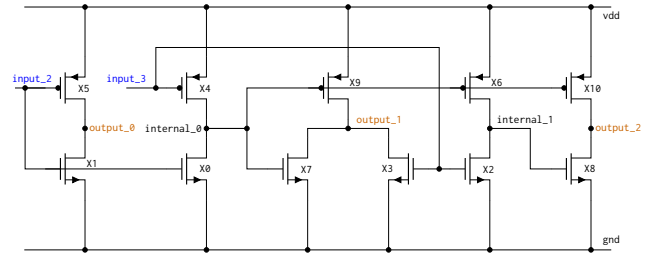
\begin{figure}
    \centering
    \resizebox{\linewidth}{!}{\begin{circuitikz}[american]
\ctikzset{tripoles/mos style=arrows}

\draw (0,6) -- (14.5,6) node[below]{\small\texttt{vdd}};
\draw (0,0)  -- (14.5,0)  node[above]{\small\texttt{gnd}};

\node[pmos] (M5)  at (1.5,4)  {};
\node[nmos] (M1)  at (1.5,2)  {};

\node[pmos] (M4)  at (4.5,4)  {};
\node[nmos] (M0)  at (4.5,2)  {};

\node[pmos] (M9)  at (8,4) {};
\node[nmos] (M7)  at (7,2) {};
\node[nmos, xscale=-1] (M3)  at (9,2) {};

\node[pmos] (M6)  at (11.5,4)  {};
\node[nmos] (M2)  at (11.5,2)  {};

\node[pmos] (M10) at (14,4) {};
\node[nmos] (M8)  at (14,2)  {};

\node[right=-8pt] at (M4)  {\small\texttt{X4}};
\node[right=-8pt] at (M0)  {\small\texttt{X0}};

\node[right=-8pt] at (M5)  {\small\texttt{X5}};
\node[xshift=5pt, yshift=4pt] at (M1)  {\small\texttt{X1}};

\node[xshift=5pt, yshift=4pt] at (M6)  {\small\texttt{X6}};
\node[right=-8pt] at (M2)  {\small\texttt{X2}};

\node[xshift=5pt, yshift=4pt] at (M9)  {\small\texttt{X9}};
\node[right=-8pt] at (M3)  {\small\texttt{X3}};
\node[right=-8pt] at (M7)  {\small\texttt{X7}};

\node[right=-8pt] at (M10) {\small\texttt{X10}};
\node[right=-8pt] at (M8)  {\small\texttt{X8}};

\foreach \p in {M4,M5,M6,M9,M10} {
  \draw (\p.S) -- (\p.S |- 0,6);
}
\foreach \n in {M0,M1,M2,M3,M7,M8} {
  \draw (\n.S) -- (\n.S |- 0,0);
}

\coordinate (nout0) at (1.5,3);
\coordinate (nint0) at (4.5,3);
\coordinate (nout1) at (8,3);
\coordinate (nint1) at (11.5,3);
\coordinate (nout2) at (14,3);

\draw (M4.D) -- (nint0) -- (M0.D);
\draw (M5.D) -- (nout0) -- (M1.D);
\draw (M6.D) -- (nint1) -- (M2.D);
\draw (M9.D) -- (nout1);
\draw (M3.D) |- (nout1);
\draw (M7.D) |- (nout1);
\draw (M10.D) -- (nout2) -- (M8.D);

\coordinate (nint0a) at (5.5,3);
\coordinate (nint1a) at (12.5,3);

\draw (M1.G) -- ++(-0.5,0) -- ++(0.0,+2.0) -- (M5.G);
\draw (M1.G) -- (M0.G);

\draw (nint0) -- (nint0a) |- (M7.G);
\draw (nint0) -- (nint0a) |- (M9.G) -- (M6.G) -- (M10.G);

\draw (nint1) -- (nint1a) |- (M8.G);

\draw (M3.G) -- (M2.G);

\coordinate (ninp3a) at (10.25,2);
\coordinate (ninp3b) at (10.25,5);
\coordinate (ninp3c) at (3.25,5);
\coordinate (ninp3d) at (3.25,4);

\draw (ninp3a) -- (ninp3b) -- (ninp3c) -- (ninp3d);

\node[circ] at (nint0) {};
\node[circ] at (nint0a) {};
\node[circ] at (nout0) {};
\node[circ] at (nint1) {};
\node[circ] at (nout1) {};
\node[circ] at (nout2) {};

\node[circ] at (1.5,0) {};
\node[circ] at (1.5,6) {};
\node[circ] at (4.5,0) {};
\node[circ] at (4.5,6) {};
\node[circ] at (7,0) {};
\node[circ] at (9,0) {};
\node[circ] at (8,6) {};
\node[circ] at (11.5,0) {};
\node[circ] at (11.5,6) {};
\node[circ] at (14,0) {};
\node[circ] at (14,6) {};

\node[circ] at ([xshift=-0.5cm]M5.G) {};

\node[circ] at (ninp3a) {};
\node[circ] at (ninp3d) {};

\node[left] at (nint0) {\small\texttt{internal\_0}};
\node[right] at (nout0) {\small\textcolor{darkorange}{\texttt{output\_0}}};
\node[xshift=25pt,yshift=6pt] at (nint1) {\small\texttt{internal\_1}};
\node[xshift=22pt,yshift=6pt] at (nout1) {\small\textcolor{darkorange}{\texttt{output\_1}}};
\node[right] at (nout2) {\small\textcolor{darkorange}{\texttt{output\_2}}};

\draw (M4.G)  -- ++(-1.0,0) node[above]{\small\textcolor{blue}{\texttt{input\_3}}};
\draw (M5.G)  -- ++(-0.75,0) node[xshift=12pt,yshift=6pt]{\small\textcolor{blue}{\texttt{input\_2}}};

\end{circuitikz}}
    \vspace{-0.6cm}
    \caption{Schematic of the best validation circuit found by \llmspm{}. The circuit uses only \texttt{input\_2} and \texttt{input\_3}, corresponding to petal length and petal width, respectively. The output nets \texttt{output\_0}, \texttt{output\_1}, and \texttt{output\_2} correspond to setosa, versicolor, and virginica.}
    \label{fig:schematic}
    \vspace{-0.5cm}
\end{figure}

Finally, Fig.~\ref{fig:schematic} shows the schematic of the best circuit on the validation set found by \llmspm{}. The complete netlist, including transistor sizes, is shown in Fig.~\ref{fig:netlist} in Appx.~\ref{appx:netlist}. Because of the size penalty in our reward, the resulting circuit is compact while still performing well on the classification task. Interestingly, the best solution in Fig.~\ref{fig:schematic} uses only two of the four available inputs, namely \emph{petal length} and \emph{petal width}. This is consistent with prior feature-contribution analyses on the Iris dataset, which identify these two attributes as the dominant contributors to the classification decision~\cite{palczewska2013interpreting}. It is notable that the synthesized circuit discovers this input subset implicitly and achieves a train accuracy of $93.4\%$, a validation accuracy of $88.4\%$, and a test accuracy of $85.9\%$, as discussed in more detail in the next section. To better understand the circuit behavior, Fig.~\ref{fig:waveforms} in Appx.~\ref{appx:output_waveforms} shows the output waveforms for the \texttt{tt} corner, with the transients for the three shuffled versions of the test split overlaid. The circuit behaves relatively ``statically,'' which is advantageous because its response does not depend strongly on the cycle at which the samples are presented. In addition, the outputs are clearly separated, indicating a large output separation and thus robustness to noise. For completeness, Fig.~\ref{fig:waveforms_corners} in Appx.~\ref{appx:output_waveforms} shows the corresponding waveforms overlaid across all corners and all shuffles.

Fig.~\ref{fig:other_good_circuits} in Appx.~\ref{appx:netlist} presents three additional strong circuits found by \llmspm{}. Comparing them shows that our method discovers diverse solutions with substantially different topologies while maintaining good performance. Since this is a non-standard synthesis task for which memorized circuit templates are unlikely to exist, the strong performance of \llmspm{} suggests that the approach can generalize to novel tasks, which is important for practical analog design.

\squeezeSubSection\subsection{Accuracy and Robustness}
To place the performance of the synthesized circuits into context, we compare them with two standard machine learning baselines implemented in \texttt{scikit-learn}~\cite{pedregosa2011scikit}:
\begin{itemize}[noitemsep, topsep=0pt, leftmargin=*]
    \item a regularized logistic regression classifier, implemented with \texttt{LogisticRegression}, and
    \item a single-hidden-layer neural network with four hidden units and a batch size of eight implemented with \texttt{MLPClassifier}.
\end{itemize}
The best models are selected on the validation set and then evaluated once on the held-out test set.

For evaluation, we study robustness to perturbations of the input voltages. Specifically, we add Gaussian noise
$\mathcal{N}(0, \sigma_\text{noise}^2 \mathbf{I})$
to each input sample $\mathbf{x}_i \in \mathbb{R}^4$ with
$\sigma_\text{noise} \in \{0\,\text{V}, 0.1\,\text{V}, 0.2\,\text{V}, 0.3\,\text{V}, 0.4\,\text{V}, 0.5\,\text{V}\}$.
For each noise level, we report results averaged over 16 independent noise realizations. This setup models practical input distortions, such as sensor noise, which can lead to slightly perturbed voltage levels. A practically useful circuit should therefore be robust to such variations.

The results are shown as boxplots in Fig.~\ref{fig:comparison_classifiers} in Appx.~\ref{appx:test_accuracies} for both the nominal \texttt{tt} corner and the multi-corner setting. In addition, Tab.~\ref{tab:test_noise_results} in Appx.~\ref{appx:test_accuracies} reports the corresponding test accuracies. Overall, the synthesized circuits show a degradation trend comparable to the ML baselines. In the nominal \texttt{tt} setting, the two best circuits are competitive with the baselines and achieve higher median accuracy at several larger noise levels. We hypothesize that this is related to optimizing the reward across multiple corners during synthesis.

\squeezeSubSection\subsection{Ablation: Prompting Strategy}
\label{subsec:prompt:sec:results}
We explored the prompt templates introduced in Sec.~\ref{subsec:llm_proposal_operator:sec:llm_spicemixer} to understand how the LLM query affects synthesis. In addition to fixed templates, we evaluated an alternating scheme that switches between the ``raw''-style and ``diff''-style prompts.

The results over nine runs using Gemma3 12B are shown in Tab.~\ref{tab:prompt} in Appx.~\ref{appx:ablation_results}. Overall, prompt design measurably affects the final reward. Variants with explicit reasoning tend to achieve higher median reward than those without reasoning. Comparing the two output formats, the ``diff''-style prompt is slightly more robust than the ``raw''-style prompt, especially without reasoning. This is consistent with edit-based outputs encouraging local modifications of elite circuits, which is advantageous for genetic algorithms. Among the evaluated settings, the alternating scheme with reasoning achieves the highest median reward. These results suggest that IGEL is sensitive to how the generation task is framed, and that better prompt design may further improve search performance.

\squeezeSubSection\subsection{Ablation: Decoding Settings}
\label{subsec:decoding:sec:results}
We also explored several LLM decoding settings instead of fixing a single choice a priori. In particular, we varied the softmax temperature $T$ and the top-$p$ probability $p_\text{top}$~\cite{DBLP:conf/iclr/HoltzmanBDFC20} to cover a range from nearly deterministic decoding to more diverse sampling. These settings control the trade-off between output stability and diversity and may therefore affect proposal quality.

The results over nine runs using Gemma3 12B are shown in Tab.~\ref{tab:decoding} in Appx.~\ref{appx:ablation_results}. We observe some variation across settings, but less than for the prompt templates. The balanced setting with $T=0.7$ and $p_\text{top}=0.9$ yields the highest median reward, while the other settings remain in a similar range. Overall, IGEL does not require highly specialized decoding settings. Moderate stochasticity appears to be a reasonable default for this task.

\squeezeSubSection\subsection{Ablation: Model Choice and Size}
\label{subsec:model:sec:results}
Finally, we explored model family and size using Gemma3~\cite{gemma3_technical_report_2025} and Qwen3.5~\cite{qwen3_5_2026}. The results are reported in Tab.~\ref{tab:reward_models} in Appx.~\ref{appx:ablation_results}.

Overall, performance depends on both model family and size. Within Gemma3, larger models generally perform better, although not strictly monotonically for every statistic. Across all models, Qwen3.5 27B performs best, reaching the highest median and maximum reward. Compared with Gemma3 27B, it improves the median reward by $+0.05$. We attribute this to overall model quality, reflecting the advantages of a newer model with stronger benchmark performance. We therefore use Qwen3.5 27B as the IGEL model in the main comparison.

To better understand model differences, we also analyzed response lengths in Tab.~\ref{tab:response_length_stats} in Appx.~\ref{appx:response_length_statistics}. Gemma3 produces relatively short outputs, whereas Qwen generates substantially longer responses, partly due to explicit thinking tokens. In particular, Qwen3.5 27B generates much longer responses than the other models. This may indicate that additional test-time reasoning is useful, although our results show only correlation, not causation.

\squeezeSubSection\subsection{Ablation: Operator Mixture}
\label{subsec:igel_only:sec:results}
We further ablate whether IGEL should replace the conventional \spm{} operators or complement them. We compare the default operator mixture against an IGEL-only variant without crossover, mutation, or pruning. The IGEL-only variant runs for $18{,}816$ proposal steps, matching the LLM-call budget of one full \llmspm{} run. Both settings use the same reasoning-enabled alternating ``raw''/``diff'' prompts with $K=3$ inspirations.

The results are reported in Tab.~\ref{tab:igel_only} in Appx.~\ref{appx:ablation_results}. IGEL alone performs substantially worse than the operator mixture. At the same proposal budget of $18{,}816$ steps, the median reward decreases from $0.689$ to $0.619$. The gap is even larger relative to the longer $131{,}072$-step operator-mixture run with the same LLM-call budget, whose median reward is $0.763$. Notably, the best IGEL-only run ($0.656$) remains below the worst operator-mixture run at the same proposal budget ($0.665$). These results show that IGEL is most effective as a complementary proposal operator: it injects useful structured variations, while conventional genetic operators remain important for local refinement, recombination, and search dynamics.

\squeezeSection\section{Conclusions and Outlook}
\label{sec:conclusions_and_outlook}

We introduced \llmspm{}, a hybrid analog circuit synthesis method that augments \spm{} with IGEL, an LLM-based proposal operator. Rather than asking the LLM to generate circuits from scratch, we use it to propose new candidate netlists from elite solutions within a genetic search loop. This enables the LLM to contribute structured variations, while SPICE simulation remains the source of truth for evaluating circuit quality.

We evaluated the method on analog discriminant-function synthesis for the Iris classification task. The results show that IGEL improves search performance over \spm{}, reduces premature convergence, and yields compact transistor-level circuits with strong performance. Ablations show that prompt design and model choice are important in this setting, while the decoding configuration has a comparatively smaller effect.

Several directions remain for future work. First, the gap between training and validation/test reward suggests that improved reward design or explicit regularization could further enhance generalization. Second, robustness could be strengthened by incorporating noisy or perturbed inputs directly during synthesis rather than only during post-training evaluation. Third, it would be interesting to explore richer LLM-based proposal strategies, such as verbalized sampling~\cite{zhang2025verbalized}, self-reflective refinement~\cite{madaan2023self,ye2024reevo}, or multi-stage proposal-and-repair schemes~\cite{zhang2025analogxpert,lai2026analogcoder}.

Overall, our results suggest that for non-standard analog synthesis tasks, LLMs are most effective not as standalone designers, but as proposal generators embedded within a simulation-driven evolutionary search loop.

\bibliographystyle{IEEEtran}
\bibliography{references}

@article{uhlich2025spicemixer,
  title={SPICEMixer-Netlist-Level Circuit Evolution},
  author={Uhlich, Stefan and Bonetti, Andrea and Venkitaraman, Arun and Hsieh, Chia-Yu and Gen{\c{c}}er, Ya{\u{g}}{\i}z and G{\"u}rsoy, Mustafa Emre and Matsuo, Ryoga and Servadei, Lorenzo},
  journal={arXiv preprint arXiv:2506.01497},
  year={2025}
}

@book{razavi2017analogcmos,
  author    = {Behzad Razavi},
  title     = {Design of Analog CMOS Integrated Circuits},
  edition   = {2},
  year      = {2017},
  publisher = {McGraw-Hill Education}
}

@article{uhlich2024graco,
  title={GraCo--A Graph Composer for Integrated Circuits},
  author={Uhlich, Stefan and Bonetti, Andrea and Venkitaraman, Arun and Momeni, Ali and Matsuo, Ryoga and Hsieh, Chia-Yu and Ohbuchi, Eisaku and Servadei, Lorenzo},
  journal={arXiv preprint arXiv:2411.13890},
  year={2024}
}

@inproceedings{vijayaraghavan2025autocircuit,
  title={AUTOCIRCUIT-RL: Reinforcement Learning-Driven LLM for Automated Circuit Topology Generation},
  author={Vijayaraghavan, Prashanth and Shi, Luyao and Degan, Ehsan and Mukherjee, Vandana and Zhang, Xin},
  booktitle={International Conference on Machine Learning},
  pages={61498--61512},
  year={2025},
  organization={PMLR}
}

@article{lyu2017efficient,
  title={An efficient Bayesian optimization approach for automated optimization of analog circuits},
  author={Lyu, Wenlong and Xue, Pan and Yang, Fan and Yan, Changhao and Hong, Zhiliang and Zeng, Xuan and Zhou, Dian},
  journal={IEEE Transactions on Circuits and Systems I: Regular Papers},
  volume={65},
  number={6},
  pages={1954--1967},
  year={2017},
  publisher={IEEE}
}

@inproceedings{gu2024tss,
  title={tss-bo: Scalable bayesian optimization for analog circuit sizing via truncated subspace sampling},
  author={Gu, Tianchen and Wang, Jiaqi and Bi, Zhaori and Yan, Changhao and Yang, Fan and Qin, Yajie and Cui, Tao and Zeng, Xuan},
  booktitle={2024 Design, Automation \& Test in Europe Conference \& Exhibition (DATE)},
  pages={1--6},
  year={2024},
  organization={IEEE}
}

@inproceedings{settaluri2020autockt,
  title={AutoCkt: deep reinforcement learning of analog circuit designs},
  author={Settaluri, Keertana and Haj-Ali, Ameer and Huang, Qijing and Hakhamaneshi, Kourosh and Nikolic, Borivoje},
  booktitle={Proceedings of the 23rd Conference on Design, Automation and Test in Europe},
  pages={490--495},
  year={2020}
}

@inproceedings{wang2020gcn,
  title={GCN-RL circuit designer: Transferable transistor sizing with graph neural networks and reinforcement learning},
  author={Wang, Hanrui and Wang, Kuan and Yang, Jiacheng and Shen, Linxiao and Sun, Nan and Lee, Hae-Seung and Han, Song},
  booktitle={2020 57th ACM/IEEE Design Automation Conference (DAC)},
  pages={1--6},
  year={2020},
  organization={IEEE}
}

@inproceedings{budak2021dnn,
  title={DNN-Opt: An RL Inspired Optimization for Analog Circuit Sizing Using Deep Neural Networks},
  author={Budak, Ahmet F and Bhansali, Prateek and Liu, Bo and Sun, Nan and Pan, David Z and Kashyap, Chandramouli V},
  booktitle={Proceedings of the 58th Annual ACM/IEEE Design Automation Conference},
  pages={1219--1224},
  year={2021}
}

@inproceedings{kim2025ppaas,
  title={Ppaas: Pvt and pareto aware analog sizing via goal-conditioned reinforcement learning},
  author={Kim, Seunggeun and Wang, Ziyi and Lee, Sungyoung and Oh, Youngmin and Zhu, Hanqing and Kim, Doyun and Pan, David Z},
  booktitle={2025 IEEE/ACM International Conference On Computer Aided Design (ICCAD)},
  pages={1--9},
  year={2025},
  organization={IEEE}
}

@article{somayaji2025llm,
  title={LLM-USO: Large Language Model-based Universal Sizing Optimizer},
  author={Somayaji, NS Karthik and Li, Peng},
  journal={IEEE Transactions on Computer-Aided Design of Integrated Circuits and Systems},
  year={2025},
  publisher={IEEE}
}

@article{liu2025eesizer,
  title={Eesizer: Llm-based ai agent for sizing of analog and mixed signal circuit},
  author={Liu, Chang and Chitnis, Danial},
  journal={IEEE Transactions on Circuits and Systems I: Regular Papers},
  year={2025},
  publisher={IEEE}
}

@inproceedings{ahmadzadeh2025anaflow,
  title={AnaFlow: Agentic LLM-based workflow for reasoning-driven explainable and sample-efficient analog circuit sizing},
  author={Ahmadzadeh, Mohsen and Chen, Kaichang and Gielen, Georges},
  booktitle={2025 IEEE/ACM International Conference On Computer Aided Design (ICCAD)},
  pages={1--7},
  year={2025},
  organization={IEEE}
}

@inproceedings{lu2022topology,
  title={Topology optimization of operational amplifier in continuous space via graph embedding},
  author={Lu, Jialin and Lei, Liangbo and Yang, Fan and Shang, Li and Zeng, Xuan},
  booktitle={2022 Design, Automation \& Test in Europe Conference \& Exhibition (DATE)},
  pages={142--147},
  year={2022},
  organization={IEEE}
}

@article{zhao2022analog,
  title={Analog integrated circuit topology synthesis with deep reinforcement learning},
  author={Zhao, Zhenxin and Zhang, Lihong},
  journal={IEEE Transactions on Computer-Aided Design of Integrated Circuits and Systems},
  volume={41},
  number={12},
  pages={5138--5151},
  year={2022},
  publisher={IEEE}
}

@inproceedings{chen2023total,
  title={Total: Topology optimization of operational amplifier via reinforcement learning},
  author={Chen, Zihao and Meng, Songlei and Yang, Fan and Shang, Li and Zeng, Xuan},
  booktitle={2023 24th International Symposium on Quality Electronic Design (ISQED)},
  pages={1--8},
  year={2023},
  organization={IEEE}
}

@inproceedings{poddar2024data,
  title={A data-driven analog circuit synthesizer with automatic topology selection and sizing},
  author={Poddar, Souradip and Budak, Ahmet and Zhao, Linran and Hsu, Chen-Hao and Maji, Supriyo and Zhu, Keren and Jia, Yaoyao and Pan, David Z},
  booktitle={2024 Design, Automation \& Test in Europe Conference \& Exhibition (DATE)},
  pages={1--6},
  year={2024},
  organization={IEEE}
}

@inproceedings{shen2025into,
  title={INTO-OA: Interpretable Topology Optimization for Operational Amplifiers},
  author={Shen, Jinyi and Yang, Fan and Shang, Li and Bi, Zhaori and Yan, Changhao and Zhou, Dian and Zeng, Xuan},
  booktitle={2025 Design, Automation \& Test in Europe Conference (DATE)},
  pages={1--7},
  year={2025},
  organization={IEEE}
}

@incollection{koza1996automated,
  title={Automated design of both the topology and sizing of analog electrical circuits using genetic programming},
  author={Koza, John R and Bennett III, Forrest H and Andre, David and Keane, Martin A},
  booktitle={Artificial intelligence in design’96},
  pages={151--170},
  year={1996},
  publisher={Springer}
}

@inproceedings{hu2005open,
  title={Open-ended robust design of analog filters using genetic programming},
  author={Hu, Jianjun and Zhong, Xiwei and Goodman, Erik D},
  booktitle={Proceedings of the 7th annual conference on Genetic and evolutionary computation},
  pages={1619--1626},
  year={2005}
}

@article{rojec2018analog,
  title={Analog circuit topology representation for automated synthesis and optimization},
  author={Rojec, {\v{Z}}iga and Olen{\v{s}}ek, Jernej and Fajfar, Iztok},
  journal={Informacije MIDEM},
  volume={48},
  number={1},
  pages={29--40},
  year={2018}
}

@inproceedings{fan2021specification,
  title={From specification to topology: Automatic power converter design via reinforcement learning},
  author={Fan, Shaoze and Cao, Ningyuan and Zhang, Shun and Li, Jing and Guo, Xiaoxiao and Zhang, Xin},
  booktitle={2021 IEEE/ACM International Conference On Computer Aided Design (ICCAD)},
  pages={1--9},
  year={2021},
  organization={IEEE}
}

@article{gao2026powergenie,
  title={PowerGenie: Analytically-Guided Evolutionary Discovery of Superior Reconfigurable Power Converters},
  author={Gao, Jian and Zou, Yiwei and Pradhan, Abhishek and Huang, Wenhao and Su, Yumin and Yang, Kaiyuan and Zhang, Xuan},
  journal={arXiv preprint arXiv:2601.21984},
  year={2026}
}

@article{gao2025analoggenie,
  title={Analoggenie: A generative engine for automatic discovery of analog circuit topologies},
  author={Gao, Jian and Cao, Weidong and Yang, Junyi and Zhang, Xuan},
  journal={arXiv preprint arXiv:2503.00205},
  year={2025}
}

@article{ren2026spicefuzz,
  title={SpiceFuzz: LLM-Based Fuzzing for Spice Circuit Simulator Tools Bug Detection},
  author={Ren, Zhilei and Liu, Huijiang and Guo, Shikai and Guo, Yi and Li, Xiaochen and Jiang, He},
  journal={ACM Transactions on Design Automation of Electronic Systems},
  year={2026},
  publisher={ACM New York, NY}
}

@article{kim2026analogtobi,
  title={AnalogToBi: Device-Level Analog Circuit Topology Generation via Bipartite Graph and Grammar Guided Decoding},
  author={Kim, Seungmin and Kim, Mingun and Lee, Yuna and Kim, Yulhwa},
  journal={arXiv preprint arXiv:2603.08720},
  year={2026}
}

@inproceedings{lai2025analogcoder,
  title={Analogcoder: Analog circuit design via training-free code generation},
  author={Lai, Yao and Lee, Sungyoung and Chen, Guojin and Poddar, Souradip and Hu, Mengkang and Pan, David Z and Luo, Ping},
  booktitle={Proceedings of the AAAI Conference on Artificial Intelligence},
  volume={39},
  number={1},
  pages={379--387},
  year={2025}
}

@article{lai2026analogcoder,
  title={Analogcoder-pro: Unifying analog circuit generation and optimization via multi-modal llms},
  author={Lai, Yao and Poddar, Souradip and Lee, Sungyoung and Chen, Guojin and Hu, Mengkang and Yu, Bei and Luo, Ping and Pan, David Z},
  journal={IEEE Transactions on Computer-Aided Design of Integrated Circuits and Systems},
  year={2026},
  publisher={IEEE}
}

@article{zhang2025verbalized,
  title={Verbalized sampling: How to mitigate mode collapse and unlock llm diversity},
  author={Zhang, Jiayi and Yu, Simon and Chong, Derek and Sicilia, Anthony and Tomz, Michael R and Manning, Christopher D and Shi, Weiyan},
  journal={arXiv preprint arXiv:2510.01171},
  year={2025}
}

@article{yun2025price,
  title={The Price of Format: Diversity Collapse in LLMs},
  author={Yun, Longfei and An, Chenyang and Wang, Zilong and Peng, Letian and Shang, Jingbo},
  journal={arXiv preprint arXiv:2505.18949},
  year={2025}
}

@article{wright2025epistemic,
  title={Epistemic diversity and knowledge collapse in large language models},
  author={Wright, Dustin and Masud, Sarah and Moore, Jared and Yadav, Srishti and Antoniak, Maria and Christensen, Peter Ebert and Park, Chan Young and Augenstein, Isabelle},
  journal={arXiv preprint arXiv:2510.04226},
  year={2025}
}

@misc{skywater130pdk,
  author       = {{Google and SkyWater Technology Foundry}},
  title        = {SkyWater 130nm {PDK}},
  year         = {2020},
  url          = {https://github.com/google/skywater-pdk}
}

@article{fisher1936use,
  title={The use of multiple measurements in taxonomic problems},
  author={Fisher, Ronald A},
  journal={Annals of eugenics},
  volume={7},
  number={2},
  pages={179--188},
  year={1936},
  publisher={Wiley Online Library}
}

@inproceedings{kwon2023efficient,
  title={Efficient memory management for large language model serving with pagedattention},
  author={Kwon, Woosuk and Li, Zhuohan and Zhuang, Siyuan and Sheng, Ying and Zheng, Lianmin and Yu, Cody Hao and Gonzalez, Joseph and Zhang, Hao and Stoica, Ion},
  booktitle={Proceedings of the 29th symposium on operating systems principles},
  pages={611--626},
  year={2023}
}

@inproceedings{DBLP:conf/iclr/HoltzmanBDFC20,
  author       = {Ari Holtzman and
                  Jan Buys and
                  Li Du and
                  Maxwell Forbes and
                  Yejin Choi},
  title        = {The Curious Case of Neural Text Degeneration},
  booktitle    = {8th International Conference on Learning Representations, {ICLR} 2020,
                  Addis Ababa, Ethiopia, April 26-30, 2020},
  publisher    = {OpenReview.net},
  year         = {2020},
  url          = {https://openreview.net/forum?id=rygGQyrFvH},
  bibsource    = {dblp computer science bibliography, https://dblp.org}
}

@article{salimans2017evolution,
  title={Evolution strategies as a scalable alternative to reinforcement learning},
  author={Salimans, Tim and Ho, Jonathan and Chen, Xi and Sidor, Szymon and Sutskever, Ilya},
  journal={arXiv preprint arXiv:1703.03864},
  year={2017}
}

@misc{gemma3_technical_report_2025,
  title         = {Gemma 3 Technical Report},
  author        = {{Gemma Team}},
  year          = {2025},
  eprint        = {2503.19786},
  archiveprefix = {arXiv},
  primaryclass  = {cs.CL},
  doi           = {10.48550/arXiv.2503.19786},
  url           = {https://arxiv.org/abs/2503.19786}
}

@misc{qwen3_5_2026,
  title  = {{Qwen3.5}: Towards Native Multimodal Agents},
  author = {{Qwen Team}},
  month  = feb,
  year   = {2026},
  url    = {https://qwen.ai/blog?id=qwen3.5}
}

@inproceedings{zhang2025analogxpert,
  title={Analogxpert: Automating analog topology synthesis by incorporating circuit design expertise into large language models},
  author={Zhang, Haoyi and Sun, Shizhao and Lin, Yibo and Wang, Runsheng and Bian, Jiang},
  booktitle={2025 International Symposium of Electronics Design Automation (ISEDA)},
  pages={772--777},
  year={2025},
  organization={IEEE}
}

@article{romera2024funsearch,
  title   = {Mathematical discoveries from program search with large language models},
  author  = {Romera-Paredes, Bernardino and Barekatain, Mohammadamin and Novikov, Alexander and Balog, Matej and Kumar, M. Pawan and Dupont, Emilien and Ruiz, Francisco J. R. and Ellenberg, Jordan S. and Wang, Pengming and Fawzi, Omar and Kohli, Pushmeet and Fawzi, Alhussein},
  journal = {Nature},
  volume  = {625},
  pages   = {468--475},
  year    = {2024},
  doi     = {10.1038/s41586-023-06924-6}
}

@misc{liu2023ael,
  title         = {Algorithm Evolution Using Large Language Model},
  author        = {Liu, Fei and Tong, Xialiang and Yuan, Mingxuan and Zhang, Qingfu},
  year          = {2023},
  eprint        = {2311.15249},
  archivePrefix = {arXiv},
  primaryClass  = {cs.NE},
  doi           = {10.48550/arXiv.2311.15249}
}

@misc{liu2024eoh,
  title         = {Evolution of Heuristics: Towards Efficient Automatic Algorithm Design Using Large Language Model},
  author        = {Liu, Fei and Tong, Xialiang and Yuan, Mingxuan and Lin, Xi and Luo, Fu and Wang, Zhenkun and Lu, Zhichao and Zhang, Qingfu},
  year          = {2024},
  eprint        = {2401.02051},
  archivePrefix = {arXiv},
  primaryClass  = {cs.NE},
  doi           = {10.48550/arXiv.2401.02051}
}

@misc{ye2024reevo,
  title         = {ReEvo: Large Language Models as Hyper-Heuristics with Reflective Evolution},
  author        = {Ye, Haoran and Wang, Jiarui and Cao, Zhiguang and Berto, Federico and Hua, Chuanbo and Kim, Haeyeon and Park, Jinkyoo and Song, Guojie},
  year          = {2024},
  eprint        = {2402.01145},
  archivePrefix = {arXiv},
  primaryClass  = {cs.NE},
  doi           = {10.48550/arXiv.2402.01145}
}

@misc{vanstein2024llamea,
  title         = {LLaMEA: A Large Language Model Evolutionary Algorithm for Automatically Generating Metaheuristics},
  author        = {van Stein, Niki and B{\"a}ck, Thomas},
  year          = {2024},
  eprint        = {2405.20132},
  archivePrefix = {arXiv},
  primaryClass  = {cs.NE},
  doi           = {10.48550/arXiv.2405.20132}
}

@misc{novikov2025alphaevolve,
  title         = {AlphaEvolve: A coding agent for scientific and algorithmic discovery},
  author        = {Novikov, Alexander and Vu, Ngan and Eisenberger, Marvin and Dupont, Emilien and Huang, Po-Sen and Wagner, Adam Zsolt and Shirobokov, Sergey and Kozlovskii, Borislav and Ruiz, Francisco J. R. and Mehrabian, Abbas and Kumar, M. Pawan and See, Abigail and Chaudhuri, Swarat and Holland, George and Davies, Alex and Nowozin, Sebastian and Kohli, Pushmeet and Balog, Matej},
  year          = {2025},
  eprint        = {2506.13131},
  archivePrefix = {arXiv},
  primaryClass  = {cs.AI},
  doi           = {10.48550/arXiv.2506.13131}
}

@misc{zhang2026trajectory,
  title         = {What Makes an LLM a Good Optimizer? A Trajectory Analysis of LLM-Guided Evolutionary Search},
  author        = {Zhang, Xinhao and Chen, Xi and Portet, Francois and Peyrard, Maxime},
  year          = {2026},
  eprint        = {2604.19440},
  archivePrefix = {arXiv},
  primaryClass  = {cs.CL},
  doi           = {10.48550/arXiv.2604.19440}
}

@article{madaan2023self,
  title={Self-refine: Iterative refinement with self-feedback},
  author={Madaan, Aman and Tandon, Niket and Gupta, Prakhar and Hallinan, Skyler and Gao, Luyu and Wiegreffe, Sarah and Alon, Uri and Dziri, Nouha and Prabhumoye, Shrimai and Yang, Yiming and others},
  journal={Advances in neural information processing systems},
  volume={36},
  pages={46534--46594},
  year={2023}
}

@article{pedregosa2011scikit,
  title={Scikit-learn: Machine Learning in Python},
  author={Pedregosa, Fabian and Varoquaux, Ga{\"e}l and Gramfort, Alexandre and Michel, Vincent and Thirion, Bertrand and Grisel, Olivier and Blondel, Mathieu and Prettenhofer, Peter and Weiss, Ron and Dubourg, Vincent and others},
  journal={Journal of Machine Learning Research},
  volume={12},
  pages={2825--2830},
  year={2011}
}

@incollection{palczewska2013interpreting,
  title={Interpreting random forest classification models using a feature contribution method},
  author={Palczewska, Anna and Palczewski, Jan and Marchese Robinson, Richard and Neagu, Daniel},
  booktitle={Integration of reusable systems},
  pages={193--218},
  year={2014},
  publisher={Springer}
}

@article{bao2026analogagent,
  title={Analogagent: Self-improving analog circuit design automation with llm agents},
  author={Bao, Zhixuan and Lin, Zhuoyi and Wang, Jiageng and Hu, Jinhai and Gao, Yuan and Wu, Yaoxin and Li, Xiaoli and Xu, Xun},
  journal={arXiv preprint arXiv:2603.23910},
  year={2026}
}

@manual{ngspice46,
  title        = {{Ngspice User's Manual}},
  author       = {{Ngspice Contributors}},
  organization = {{Ngspice Project}},
  edition      = {Version 46},
  year         = {2026},
  url          = {https://ngspice.sourceforge.io/docs/ngspice-manual.pdf},
  note         = {Accessed: 2026-05-08}
}

\clearpage
\appendix
\onecolumn

\squeezeSection\section{Examples of LLM Prompts and Outputs}
\label{appx:prompt_and_output}
Fig.~\ref{fig:full_style_template} and Fig.~\ref{fig:diff_style_template} on the following pages show example prompts used for IGEL together with the corresponding LLM outputs. These examples illustrate how the different prompt styles are formulated and what kinds of netlist modifications the model proposes.

\begin{figure}
\begin{minipage}[t]{0.54\textwidth}
\vspace{0pt}
\begin{prompt}[Prompt (\textcolor{no_reasoning}{w/o reasoning} or \textcolor{reasoning}{w/ reasoning})]
    \fontsize{5pt}{5pt}\selectfont
    \ttfamily
    \begin{Verbatim}[breaklines=true,breaksymbolleft={},commandchars=\\\{\}]
You are an expert CMOS analog IC designer and SPICE netlist generator. Your job is to design circuits as SPICE netlists.

Global rules (apply to EVERY reply):
* You may only instantiate the following MOSFETs:
  * M<index> <drain> <gate> <source> NMOS w=<width> l=<length>
  * M<index> <drain> <gate> <source> PMOS w=<width> l=<length>

* You must output EXACTLY ONE markdown fenced code block of the form:

  ```spice
  ...
  ```

\textcolor{no_reasoning}{* Outside that fenced code block:}
  \textcolor{no_reasoning}{* Output NOTHING. No text before it, no text after it.}
  \textcolor{no_reasoning}{* No explanations, comments, markdown.}
\textcolor{reasoning}{* Before that fenced code block:}
  \textcolor{reasoning}{* Output your thoughts: Analyse the inspirations, then come up with a strategy that you can use.}

* Inside the `spice` block:
  * Write ONE transistor instance per line.
  * Each transistor line has THREE connections: <drain> <gate> <source>.
  * Do NOT include comments or natural language.

* The transistor width w and length l are positive, real-valued values that you can optimize.

\textcolor{no_reasoning}{Reason through the design silently. Do NOT show your reasoning steps; only output the final netlist in a single `spice` block.}

Please design a circuit for the task: Analog classifier implementing a discriminant function for Iris with 4 input nets and 3 output nets.

Input nets:
- `net_input_0`
- `net_input_1`
- `net_input_2`
- `net_input_3`

Output nets:
- `net_output_0`
- `net_output_1`
- `net_output_2`

Supply nets:
- `net_vdd`
- `net_gnd`

The circuit is represented as a netlist and is evaluated using SPICE with a reward in [-1, 1], where higher is better.

Take inspiration from the following netlists and try to improve their structure and performance. Combine or mutate useful patterns to design a better circuit. You can do this by copying, removing, or modifying transistor lines from these inspirations.

Inspiration netlist with reward 0.701:
```spice
M0 net_internal_0 net_input_2 net_internal_1 NMOS w=25 l=3
M1 net_internal_2 net_input_3 net_output_2 NMOS w=34.7 l=54.8
M2 net_output_2 net_input_3 net_output_2 NMOS w=21.1 l=54.8
M3 net_output_2 net_input_3 net_output_2 NMOS w=21.1 l=54.8
M4 net_gnd net_internal_2 net_internal_1 NMOS w=34.7 l=68.1
M5 net_gnd net_internal_2 net_internal_1 NMOS w=34.7 l=68.1
M6 net_gnd net_internal_3 net_output_0 NMOS w=21.1 l=12.9
M7 net_internal_0 net_internal_3 net_output_0 NMOS w=16.7 l=12.9
M8 net_internal_0 net_internal_3 net_output_0 NMOS w=16.7 l=12.9
M9 net_internal_0 net_internal_3 net_output_0 NMOS w=16.7 l=12.9
M10 net_internal_0 net_internal_3 net_output_0 NMOS w=16.7 l=12.9
M11 net_output_1 net_internal_2 net_output_1 NMOS w=9.7 l=41.5
```

Inspiration netlist with reward 0.703:
```spice
M0 net_internal_0 net_input_2 net_internal_1 NMOS w=25 l=3
M1 net_internal_2 net_input_3 net_output_2 NMOS w=34.7 l=54.8
M2 net_output_2 net_input_3 net_output_2 NMOS w=21.1 l=54.8
M3 net_gnd net_internal_2 net_internal_1 NMOS w=34.7 l=68.1
M4 net_internal_0 net_internal_3 net_output_0 NMOS w=16.7 l=12.9
M5 net_internal_0 net_internal_3 net_output_0 NMOS w=16.7 l=12.9
M6 net_internal_1 net_internal_2 net_output_2 NMOS w=34.7 l=68.1
M7 net_output_1 net_internal_2 net_output_1 NMOS w=9.7 l=41.5
```

Inspiration netlist with reward 0.706:
```spice
M0 net_gnd net_input_3 net_internal_0 PMOS w=95.3 l=36.1
M1 net_internal_0 net_input_2 net_internal_1 NMOS w=25 l=3
M2 net_internal_1 net_input_1 net_internal_2 PMOS w=72.7 l=5.35
M3 net_internal_2 net_input_3 net_output_2 NMOS w=34.7 l=54.8
M4 net_output_2 net_input_3 net_output_2 NMOS w=21.1 l=54.8
M5 net_gnd net_internal_2 net_internal_1 NMOS w=34.7 l=68.1
M6 net_internal_0 net_internal_3 net_output_0 NMOS w=16.7 l=12.9
M7 net_internal_0 net_internal_3 net_output_0 NMOS w=16.7 l=12.9
M8 net_internal_0 net_internal_3 net_output_0 NMOS w=16.7 l=12.9
M9 net_internal_1 net_internal_2 net_output_2 NMOS w=34.7 l=68.1
M10 net_output_1 net_internal_2 net_output_1 NMOS w=9.7 l=41.5
```

Now propose a NEW netlist that is likely to achieve a higher reward, while respecting all global rules from the system prompt. The new circuit SHOULD NOT exceed the size of the inspiration netlists and have a MAXIMUM of 13 transistors or LESS. Avoid copying a netlist and only changing transistor widths and lengths but also consider topological changes. A "topology change" means any structural modification to the circuit graph, including at least one of:
* Rewire: change at least one of `drain`, `gate`, `source` of any transistor to a different net (not just `w/l`).
* Add device: insert a new transistor (new line) and connect it to existing nets and/or a new internal net.
* Remove device: delete an existing transistor (remove a line).
* Split/Merge nets: introduce a new internal net to break a previously shared node, or connect two previously distinct internal nets into one by rewiring.
* Change device type: replace an `NMOS` with a `PMOS` (or vice versa) and rewire if needed.

\textcolor{no_reasoning}{Remember: your reply must consist ONLY of a single `spice` fenced code block containing the netlist and nothing else.}
\textcolor{reasoning}{Remember: First reason, then answer with a `spice` fenced code block containing the netlist and nothing else.}
\end{Verbatim}
\end{prompt}
\end{minipage}\hfill
\begin{minipage}[t]{0.44\textwidth}
\vspace{0pt}
\begin{prompt}[LLM Answer (\textcolor{reasoning}{w/ reasoning}) and Netlist (generic SPICE components)]
    \fontsize{6pt}{6pt}\selectfont
    \ttfamily
    \begin{Verbatim}[breaklines=true,breaksymbolleft={},commandchars=\\\{\}]
\textcolor{reasoning}{Okay, let's analyze the provided inspiration netlists and devise a strategy for a potentially improved classifier. The common elements across the three are the use of internal nodes and repeated transistor structures for certain outputs, suggesting an attempt at redundancy or amplification. However, the redundancy isn't particularly efficient, and the structures are somewhat repetitive. The reward scores are quite similar, indicating that small changes might be beneficial.}

\textcolor{reasoning}{My strategy will be to simplify the circuit while retaining the core concepts of using internal nodes and multiple transistors for each output. I'll focus on reducing redundancy and introducing a different combination of NMOS and PMOS transistors to potentially enhance the discriminant function. Specifically, I'll try to incorporate a PMOS transistor to create a more balanced circuit. I'll also reduce the number of transistors slightly to see if it improves performance.}

```spice
M0 net_gnd net_input_3 net_internal_0 PMOS w=95.3 l=36.1
M1 net_internal_0 net_input_2 net_internal_1 NMOS w=25 l=3
M2 net_internal_1 net_input_1 net_internal_2 PMOS w=72.7 l=5.35
M3 net_internal_2 net_input_3 net_output_2 NMOS w=34.7 l=54.8
M4 net_internal_0 net_internal_3 net_output_0 NMOS w=16.7 l=12.9
M5 net_internal_0 net_internal_3 net_output_0 NMOS w=16.7 l=12.9
M6 net_internal_1 net_internal_2 net_output_2 NMOS w=34.7 l=68.1
M7 net_output_1 net_internal_2 net_output_1 NMOS w=9.7 l=41.5
```
\end{Verbatim}
\end{prompt}

\begin{prompt}[Extracted Netlist (Skywater-PDK): Reward = 0.720]
    \fontsize{6pt}{6pt}\selectfont
    \ttfamily
    \begin{Verbatim}[breaklines=true,breaksymbolleft={}]
X0 0 net_input_3 net_internal_0 net_supply_0
+ sky130_fd_pr__pfet_01v8 w=95.3 l=36.1
X1 net_internal_0 net_input_2 net_internal_1 0
+ sky130_fd_pr__nfet_01v8 w=25 l=3
X2 net_internal_1 net_input_1 net_internal_2 net_supply_0
+ sky130_fd_pr__pfet_01v8 w=72.7 l=5.35
X3 net_internal_2 net_input_3 net_output_2 0
+ sky130_fd_pr__nfet_01v8 w=34.7 l=54.8
X4 net_internal_0 net_internal_3 net_output_0 0
+ sky130_fd_pr__nfet_01v8 w=16.7 l=12.9
X5 net_internal_0 net_internal_3 net_output_0 0
+ sky130_fd_pr__nfet_01v8 w=16.7 l=12.9
X6 net_internal_1 net_internal_2 net_output_2 0
+ sky130_fd_pr__nfet_01v8 w=34.7 l=68.1
X7 net_output_1 net_internal_2 net_output_1 0
+ sky130_fd_pr__nfet_01v8 w=9.7 l=41.5
\end{Verbatim}
\end{prompt}
\end{minipage}
\caption{LLM prompt and answer for the ``raw''-style template version.}
\label{fig:full_style_template}
\end{figure}
\begin{figure}
\begin{minipage}[t]{0.54\textwidth}
\vspace{0pt}
\begin{prompt}[Prompt (\textcolor{no_reasoning}{w/o reasoning} or \textcolor{reasoning}{w/ reasoning})]
    \fontsize{5pt}{5pt}\selectfont
    \ttfamily\hspace*{1cm}
    \begin{Verbatim}[breaklines=true,breaksymbolleft={},commandchars=\\\{\}]
You are an expert CMOS analog IC designer and SPICE netlist editor. Your job is to improve netlists by proposing edits in a unified diff-style format.

Global rules (apply to EVERY reply):
* You may only introduce or keep transistor instances using the following MOSFETs:
  * M<index> <drain> <gate> <source> NMOS w=<width> l=<length>
  * M<index> <drain> <gate> <source> PMOS w=<width> l=<length>

* You must output \textcolor{no_reasoning}{EXACTLY ONE}/\textcolor{reasoning}{a} markdown fenced code block of the form:

  ```diff
  ...
  ```

\textcolor{no_reasoning}{Outside that fenced code block:}
  \textcolor{no_reasoning}{* Output NOTHING. No text before it, no text after it.}
  \textcolor{no_reasoning}{* No explanations, comments, markdown.}
\textcolor{reasoning}{* Before that fenced code block:}
  \textcolor{reasoning}{* Output your thoughts: Analyse the base netlist and the inspirations, then come up with a strategy that you can use.}

* Inside the `diff` block:
  * Each line must start with either "- " (to remove a transistor) or "+ " (to add a transistor).
  * After the "- " or "+ " prefix, write a full transistor instance line.
  * Do NOT include comments or natural language.

* The transistor width w and length l are positive, real-valued values that you can optimize.

\textcolor{no_reasoning}{Reason through the design silently. Do NOT show your reasoning steps; only output the diff in a single `diff` fenced code block.}

Please improve a circuit for the task: Analog classifier implementing a discriminant function for Iris with 4 input nets and 3 output nets.

Input nets:
- `net_input_0`
- `net_input_1`
- `net_input_2`
- `net_input_3`

Output nets:
- `net_output_0`
- `net_output_1`
- `net_output_2`

Supply nets:
- `net_vdd`
- `net_gnd`

The circuit is represented as a netlist and is evaluated using SPICE with a reward in [-1, 1], where higher is better.

Your job is to modify \textcolor{no_reasoning}{ONLY} a given base netlist by proposing a diff that is likely to improve the reward, while respecting all global rules from the system prompt.

The diff must only change transistor instance lines (M...). You may add, remove, or modify transistors by using "- " and "+ " lines.

Take inspiration from the following netlists and try to improve their structure and performance. Combine or mutate useful patterns to design a better circuit. You can do this by copying, removing, or modifying transistor lines from these inspirations.

Inspiration netlist with reward 0.765:
```spice
M0 net_internal_0 net_input_1 net_output_0 PMOS w=0.691 l=22.2
M1 net_internal_0 net_input_3 net_output_2 NMOS w=2.58 l=90.4
M2 net_internal_1 net_input_2 net_output_2 NMOS w=0.766 l=0.469
M3 net_internal_1 net_internal_0 net_output_1 NMOS w=12.2 l=67.6
M4 net_internal_1 net_internal_2 net_output_1 PMOS w=6.3 l=80.8
M5 net_internal_1 net_internal_2 net_output_2 NMOS w=6.3 l=90.4
M6 net_output_0 net_internal_2 net_output_2 NMOS w=22.9 l=90.4
```

Inspiration netlist with reward 0.768:
```spice
M0 net_internal_0 net_input_1 net_output_0 PMOS w=0.591 l=22.2
M1 net_internal_0 net_input_1 net_output_0 PMOS w=0.766 l=22.2
M2 net_internal_0 net_input_3 net_output_2 NMOS w=2.58 l=90.4
M3 net_internal_1 net_input_2 net_output_2 NMOS w=0.766 l=0.469
M4 net_internal_1 net_internal_0 net_output_1 NMOS w=12.2 l=67.6
M5 net_internal_1 net_internal_2 net_output_1 PMOS w=6.3 l=80.8
M6 net_internal_1 net_internal_2 net_output_2 NMOS w=6.3 l=74.8
M7 net_output_0 net_internal_2 net_output_2 NMOS w=22.9 l=90.4
M8 net_output_0 net_internal_2 net_output_2 NMOS w=22.9 l=90.4
```

Base netlist to modify (this is the one you must improve using a diff):
Base netlist with reward 0.767:
```spice
M0 net_internal_0 net_input_1 net_output_0 PMOS w=0.691 l=22.2
M1 net_internal_0 net_input_1 net_output_0 PMOS w=0.766 l=22.2
M2 net_internal_0 net_input_3 net_output_2 NMOS w=2.58 l=90.4
M3 net_internal_1 net_input_2 net_output_2 NMOS w=0.766 l=0.469
M4 net_internal_1 net_internal_0 net_output_1 NMOS w=12.2 l=67.6
M5 net_internal_1 net_internal_2 net_output_1 PMOS w=6.3 l=80.8
M6 net_internal_1 net_internal_2 net_output_2 NMOS w=6.3 l=90.4
M7 net_output_0 net_internal_2 net_output_2 NMOS w=22.9 l=90.4
```

Now propose a NEW set of edits to this base netlist that is likely to achieve a higher reward. Here is an example that shows how to modify a transistor line:
```diff
- M6 net_internal_1 net_internal_2 net_output_2 NMOS w=6.3 l=90.4
+ M6 net_internal_1 net_input_0 net_output_2 NMOS w=6.3 l=90.4
```

Return \textcolor{no_reasoning}{ONLY} a unified diff that transforms the base netlist into your improved netlist, following these rules:
* Use "- " lines to indicate transistors that should be removed from the base netlist.
* Use "+ " lines to indicate new or modified transistors to be present in the final netlist.
* Do NOT include any unchanged lines.

\textcolor{no_reasoning}{Remember: your reply must consist ONLY of a single `diff` fenced code block containing the diff and nothing else.} Avoid only changing transistor widths and lengths but also consider topological changes. A "topology change" means any structural modification to the circuit graph, including at least one of:
* Rewire: change at least one of `drain`, `gate`, `source` of any transistor to a different net (not just `w/l`).
* Add device: insert a new transistor (new line) and connect it to existing nets and/or a new internal net.
* Remove device: delete an existing transistor (remove a line).
* Split/Merge nets: introduce a new internal net to break a previously shared node, or connect two previously distinct internal nets into one by rewiring.
* Change device type: replace an `NMOS` with a `PMOS` (or vice versa) and rewire if needed.

\textcolor{reasoning}{Remember: First reason, then reply with a `diff` fenced code block containing the diff and nothing else.}
\end{Verbatim}
\end{prompt}
\end{minipage}\hfill
\begin{minipage}[t]{0.44\textwidth}
\vspace{0pt}
\begin{prompt}[LLM Answer (\textcolor{reasoning}{w/ reasoning}) and Netlist (generic SPICE components)]
    \fontsize{6pt}{6pt}\selectfont
    \ttfamily\hspace*{1cm}
    \begin{Verbatim}[breaklines=true,breaksymbolleft={},commandchars=\\\{\}]
\textcolor{reasoning}{The base netlist has a reward of 0.767. The inspiration netlists have rewards of 0.765 and 0.768. The differences between the base and the 0.768 netlist are subtle, primarily involving duplication of transistor M8. The base netlist is already quite close to the best performing netlist.}

\textcolor{reasoning}{The strategy is to duplicate transistor M7, as it is present in the best performing netlist. This introduces redundancy and might improve the classifier's robustness and accuracy.}

```diff
- M7 net_output_0 net_internal_2 net_output_2 NMOS w=22.9 l=90.4
+ M7 net_output_0 net_internal_2 net_output_2 NMOS w=22.9 l=90.4
+ M8 net_output_0 net_internal_2 net_output_2 NMOS w=22.9 l=90.4
```
\end{Verbatim}
\end{prompt}

\begin{prompt}[Extracted Netlist (Skywater-PDK): Reward = 0.770]
    \fontsize{6pt}{6pt}\selectfont
    \ttfamily\hspace*{1cm}
    \begin{Verbatim}[breaklines=true,breaksymbolleft={}]
X0 net_internal_0 net_input_1 net_output_0 net_supply_0
+ sky130_fd_pr__pfet_01v8 w=0.691 l=22.2
X1 net_internal_0 net_input_1 net_output_0 net_supply_0
+ sky130_fd_pr__pfet_01v8 w=0.766 l=22.2
X2 net_internal_0 net_input_3 net_output_2 0
+ sky130_fd_pr__nfet_01v8 w=2.58 l=90.4
X3 net_internal_1 net_input_2 net_output_2 0
+ sky130_fd_pr__nfet_01v8 w=0.766 l=0.469
X4 net_internal_1 net_internal_0 net_output_1 0
+ sky130_fd_pr__nfet_01v8 w=12.2 l=67.6
X5 net_internal_1 net_internal_2 net_output_1 net_supply_0
+ sky130_fd_pr__pfet_01v8 w=6.3 l=80.8
X6 net_internal_1 net_internal_2 net_output_2 0
+ sky130_fd_pr__nfet_01v8 w=6.3 l=90.4
X7 net_output_0 net_internal_2 net_output_2 0
+ sky130_fd_pr__nfet_01v8 w=22.9 l=90.4
X8 net_output_0 net_internal_2 net_output_2 0
+ sky130_fd_pr__nfet_01v8 w=22.9 l=90.4
\end{Verbatim}
\end{prompt}
\end{minipage}
\caption{LLM prompt and answer for the ``diff''-style template}
\label{fig:diff_style_template}
\end{figure}

\squeezeSection\section{Netlists of the Best Validation Circuits}
\label{appx:netlist}
\begin{figure}[H]
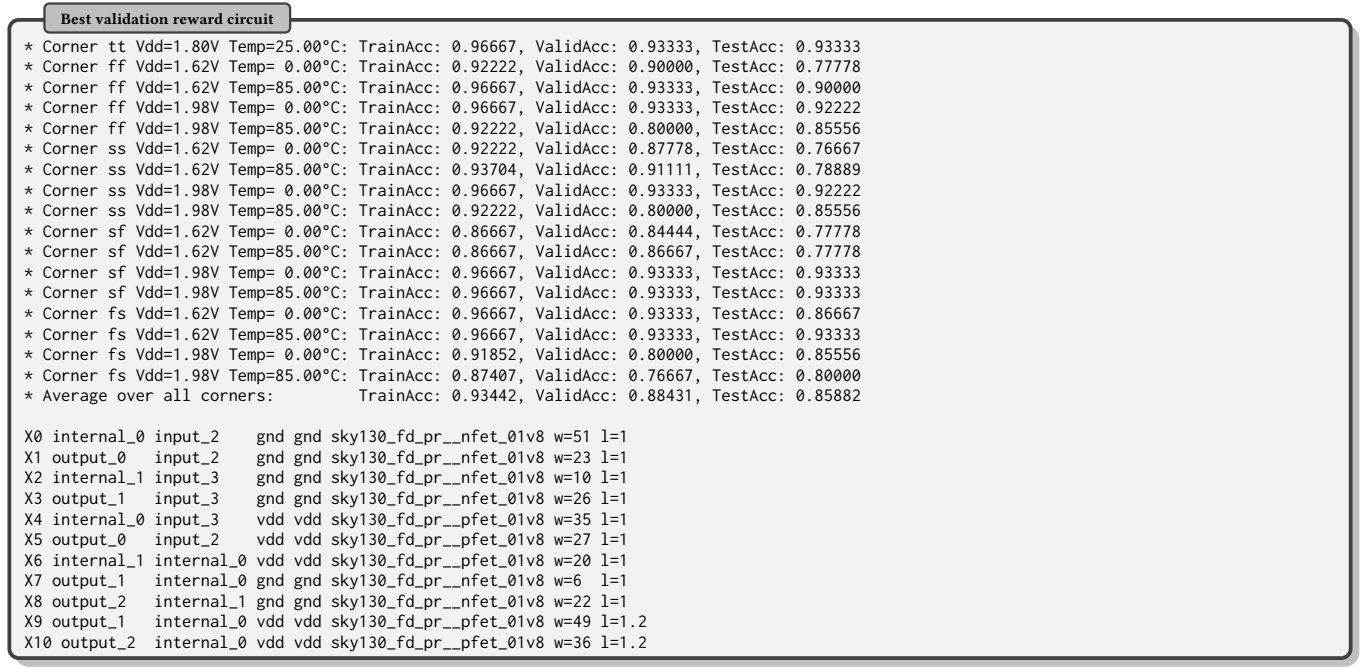

\begin{prompt}[Best validation reward circuit]
\footnotesize
\begin{Verbatim}
* Corner tt Vdd=1.80V Temp=25.00°C: TrainAcc: 0.96667, ValidAcc: 0.93333, TestAcc: 0.93333
* Corner ff Vdd=1.62V Temp= 0.00°C: TrainAcc: 0.92222, ValidAcc: 0.90000, TestAcc: 0.77778
* Corner ff Vdd=1.62V Temp=85.00°C: TrainAcc: 0.96667, ValidAcc: 0.93333, TestAcc: 0.90000
* Corner ff Vdd=1.98V Temp= 0.00°C: TrainAcc: 0.96667, ValidAcc: 0.93333, TestAcc: 0.92222
* Corner ff Vdd=1.98V Temp=85.00°C: TrainAcc: 0.92222, ValidAcc: 0.80000, TestAcc: 0.85556
* Corner ss Vdd=1.62V Temp= 0.00°C: TrainAcc: 0.92222, ValidAcc: 0.87778, TestAcc: 0.76667
* Corner ss Vdd=1.62V Temp=85.00°C: TrainAcc: 0.93704, ValidAcc: 0.91111, TestAcc: 0.78889
* Corner ss Vdd=1.98V Temp= 0.00°C: TrainAcc: 0.96667, ValidAcc: 0.93333, TestAcc: 0.92222
* Corner ss Vdd=1.98V Temp=85.00°C: TrainAcc: 0.92222, ValidAcc: 0.80000, TestAcc: 0.85556
* Corner sf Vdd=1.62V Temp= 0.00°C: TrainAcc: 0.86667, ValidAcc: 0.84444, TestAcc: 0.77778
* Corner sf Vdd=1.62V Temp=85.00°C: TrainAcc: 0.86667, ValidAcc: 0.86667, TestAcc: 0.77778
* Corner sf Vdd=1.98V Temp= 0.00°C: TrainAcc: 0.96667, ValidAcc: 0.93333, TestAcc: 0.93333
* Corner sf Vdd=1.98V Temp=85.00°C: TrainAcc: 0.96667, ValidAcc: 0.93333, TestAcc: 0.93333
* Corner fs Vdd=1.62V Temp= 0.00°C: TrainAcc: 0.96667, ValidAcc: 0.93333, TestAcc: 0.86667
* Corner fs Vdd=1.62V Temp=85.00°C: TrainAcc: 0.96667, ValidAcc: 0.93333, TestAcc: 0.93333
* Corner fs Vdd=1.98V Temp= 0.00°C: TrainAcc: 0.91852, ValidAcc: 0.80000, TestAcc: 0.85556
* Corner fs Vdd=1.98V Temp=85.00°C: TrainAcc: 0.87407, ValidAcc: 0.76667, TestAcc: 0.80000
* Average over all corners:         TrainAcc: 0.93442, ValidAcc: 0.88431, TestAcc: 0.85882

X0 internal_0 input_2    gnd gnd sky130_fd_pr__nfet_01v8 w=51 l=1
X1 output_0   input_2    gnd gnd sky130_fd_pr__nfet_01v8 w=23 l=1
X2 internal_1 input_3    gnd gnd sky130_fd_pr__nfet_01v8 w=10 l=1
X3 output_1   input_3    gnd gnd sky130_fd_pr__nfet_01v8 w=26 l=1
X4 internal_0 input_3    vdd vdd sky130_fd_pr__pfet_01v8 w=35 l=1
X5 output_0   input_2    vdd vdd sky130_fd_pr__pfet_01v8 w=27 l=1
X6 internal_1 internal_0 vdd vdd sky130_fd_pr__pfet_01v8 w=20 l=1
X7 output_1   internal_0 gnd gnd sky130_fd_pr__nfet_01v8 w=6  l=1
X8 output_2   internal_1 gnd gnd sky130_fd_pr__nfet_01v8 w=22 l=1
X9 output_1   internal_0 vdd vdd sky130_fd_pr__pfet_01v8 w=49 l=1.2
X10 output_2  internal_0 vdd vdd sky130_fd_pr__pfet_01v8 w=36 l=1.2
\end{Verbatim}
\end{prompt}
\caption{Netlist of the circuit with the best validation reward found by \llmspm{}.}
\label{fig:netlist}
\end{figure}

Fig.~\ref{fig:netlist} shows the transistor-level netlist of the circuit with the highest validation reward. Averaged over all 17 process, voltage, and temperature corners, this circuit achieves train/validation/test rewards of $0.855$/$0.811$/$0.780$. The per-corner accuracies listed above the netlist provide additional detail on its robustness across operating conditions. In addition, Fig.~\ref{fig:other_good_circuits} shows three more circuits with strong validation performance. Comparing them further illustrates the diversity of solutions discovered by \llmspm{}.

\clearpage

\begin{figure}[H]
\begin{prompt}[Second best validation reward circuit with train / validation / test reward of $0.815$ / $0.787$ / $0.760$]
\footnotesize
\begin{Verbatim}
X0 internal_0 input_2    gnd      gnd sky130_fd_pr__nfet_01v8 w=31   l=1
X1 output_0   input_2    gnd      gnd sky130_fd_pr__nfet_01v8 w=13   l=1
X2 output_1   input_3    gnd      gnd sky130_fd_pr__nfet_01v8 w=3.5  l=1
X3 internal_0 input_2    vdd      vdd sky130_fd_pr__pfet_01v8 w=3.5  l=1
X4 internal_1 input_0    vdd      gnd sky130_fd_pr__nfet_01v8 w=3.5  l=0.17
X5 output_0   input_2    vdd      vdd sky130_fd_pr__pfet_01v8 w=3.5  l=1
X6 output_2   input_3    vdd      gnd sky130_fd_pr__nfet_01v8 w=38.2 l=7
X7 output_1   internal_0 gnd      gnd sky130_fd_pr__nfet_01v8 w=2    l=1
X8 output_2   internal_1 gnd      gnd sky130_fd_pr__nfet_01v8 w=2    l=1
X9 output_0   internal_1 output_2 vdd sky130_fd_pr__pfet_01v8 w=70   l=17
X10 output_1  internal_0 vdd      vdd sky130_fd_pr__pfet_01v8 w=5    l=1
\end{Verbatim}
\end{prompt}

\vspace{0.2cm}

\begin{prompt}[Third best validation reward circuit with train / validation / test reward of $0.795$ / $0.784$ / $0.753$]
\footnotesize
\begin{Verbatim}
X0 output_1   input_1    gnd      gnd sky130_fd_pr__nfet_01v8 w=10 l=2
X1 output_2   input_1    gnd      gnd sky130_fd_pr__nfet_01v8 w=20 l=2
X2 internal_0 input_2    gnd      gnd sky130_fd_pr__nfet_01v8 w=25 l=0.211
X3 output_0   input_2    gnd      gnd sky130_fd_pr__nfet_01v8 w=20 l=2
X4 output_1   input_3    gnd      gnd sky130_fd_pr__nfet_01v8 w=10 l=2
X5 internal_0 input_2    output_0 vdd sky130_fd_pr__pfet_01v8 w=25 l=0.211
X6 output_0   input_3    vdd      vdd sky130_fd_pr__pfet_01v8 w=22 l=2
X7 output_2   input_3    vdd      gnd sky130_fd_pr__nfet_01v8 w=20 l=2
X8 output_1   internal_0 vdd      vdd sky130_fd_pr__pfet_01v8 w=20 l=2
\end{Verbatim}
\end{prompt}

\vspace{0.2cm}

\begin{prompt}[Fourth best validation reward circuit with train / validation / test reward of $0.747$ / $0.747$ / $0.702$]
\footnotesize
\begin{Verbatim}
X0 output_1   input_1    gnd      gnd sky130_fd_pr__nfet_01v8 w=10   l=1
X1 output_0   input_3    gnd      gnd sky130_fd_pr__nfet_01v8 w=80   l=1
X2 output_1   input_3    gnd      gnd sky130_fd_pr__nfet_01v8 w=2.17 l=0.42
X3 output_0   input_2    vdd      vdd sky130_fd_pr__pfet_01v8 w=3    l=1
X4 output_1   input_1    vdd      vdd sky130_fd_pr__pfet_01v8 w=3    l=1
X5 output_2   input_3    output_2 gnd sky130_fd_pr__nfet_01v8 w=1.5  l=1
X6 internal_0 internal_0 output_2 gnd sky130_fd_pr__nfet_01v8 w=1.3  l=9.35
X7 output_1   internal_0 vdd      vdd sky130_fd_pr__pfet_01v8 w=10.9 l=46.5
\end{Verbatim}
\end{prompt}
\caption{Netlists of additional circuits with strong validation performance.}
\label{fig:other_good_circuits}
\end{figure}

\clearpage

\squeezeSection\section{Output Waveforms Across Shuffles and Corners}
\label{appx:output_waveforms}
Fig.~\ref{fig:waveforms_corners} shows the input and output waveforms for the two circuits with the best validation performance. The figure includes results for all considered corners and for all three shuffled versions of the test split. This provides additional insight into how the circuit behavior varies across operating conditions.

\begin{minipage}[t]{0.47\textwidth}
\begin{figure}[H]
    \centering
    \begin{subfigure}{\linewidth}
        \centering
        \resizebox{\linewidth}{!}{\includegraphics[trim=0 10 0 0]{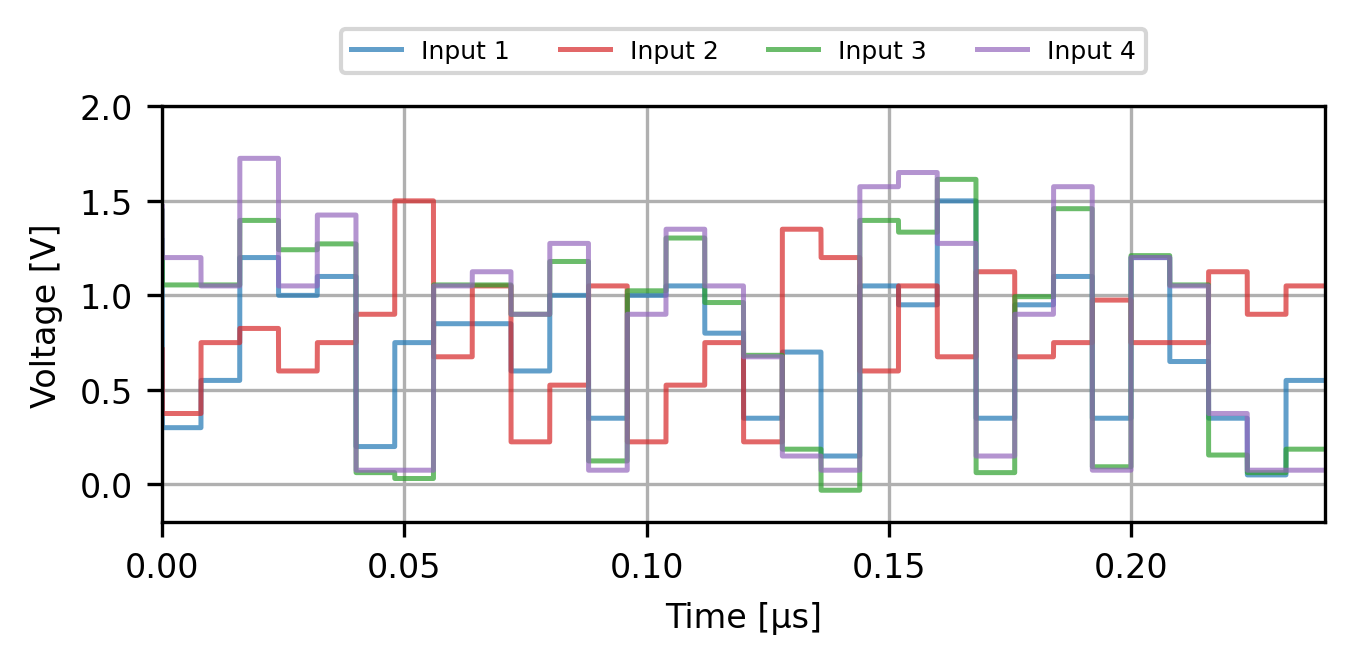}}
        \caption{Input waveforms}
    \end{subfigure}
    \begin{subfigure}{\linewidth}
        \centering
        \resizebox{\linewidth}{!}{\includegraphics[trim=0 10 0 0]{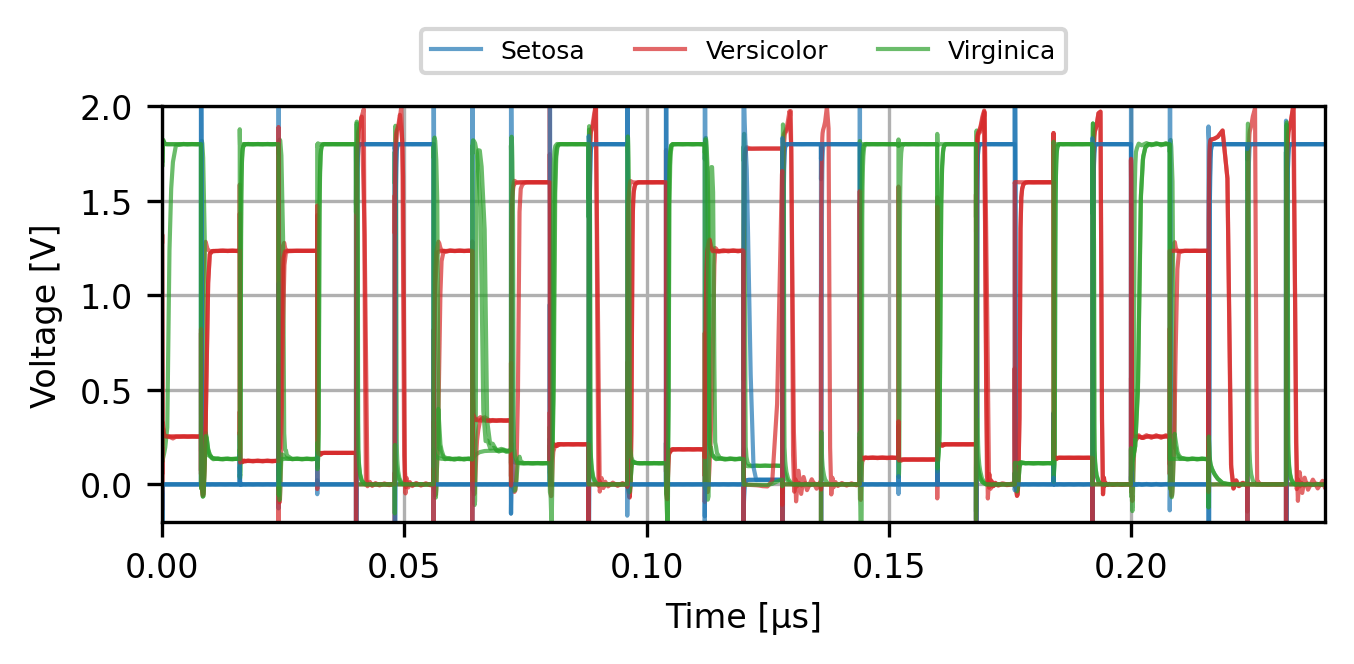}}
        \caption{Output waveforms (best validation circuit)}
    \end{subfigure}
    \begin{subfigure}{\linewidth}
        \centering
        \resizebox{\linewidth}{!}{\includegraphics[trim=0 10 0 0]{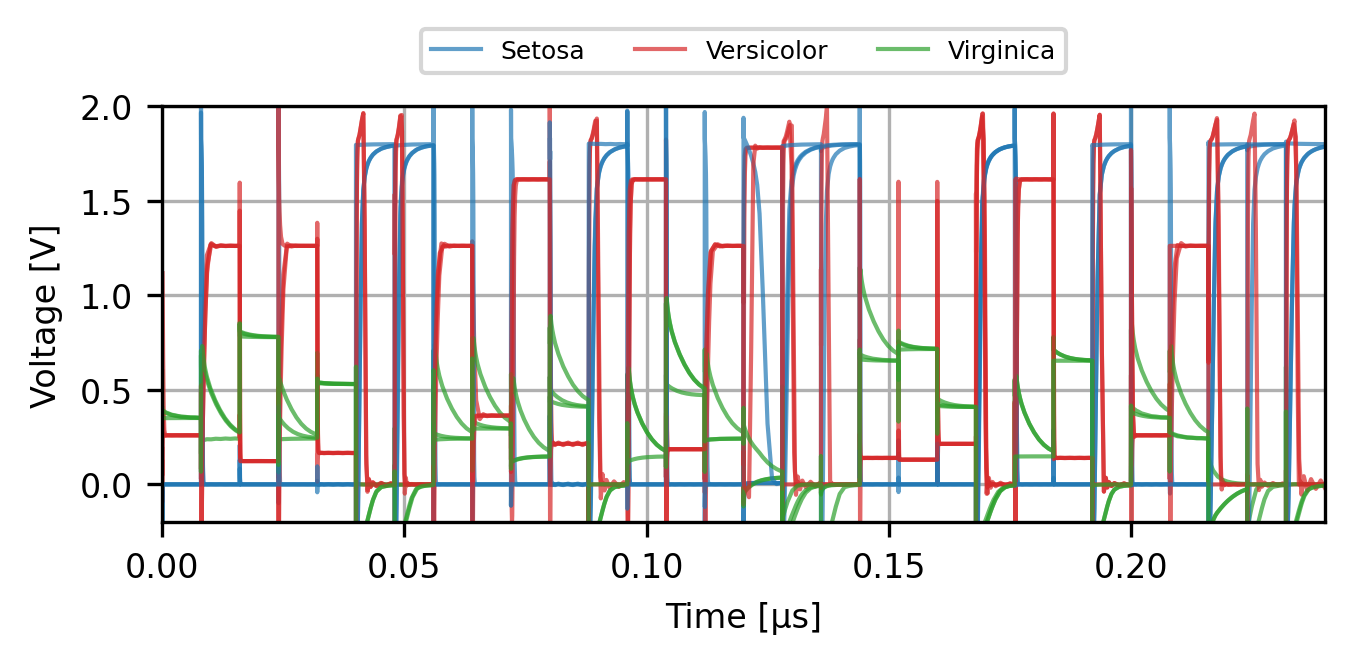}}
        \caption{Output waveforms (second-best validation circuit)}
    \end{subfigure}
    \caption{Voltage transients evaluated on the test split. The plots overlay the waveforms obtained from the three shuffled versions of the test split.}
    \label{fig:waveforms}
\end{figure}
\end{minipage}
\hfill
\begin{minipage}[t]{0.47\textwidth}
\begin{figure}[H]
    \centering
    \begin{subfigure}{\linewidth}
        \centering
        \resizebox{\linewidth}{!}{\includegraphics[trim=0 10 0 0]{figures/waveforms_overlay_shuffles_inputs.png}}
        \caption{Input waveforms}
    \end{subfigure}
    \begin{subfigure}{\linewidth}
        \centering
        \resizebox{\linewidth}{!}{\includegraphics[trim=0 10 0 0]{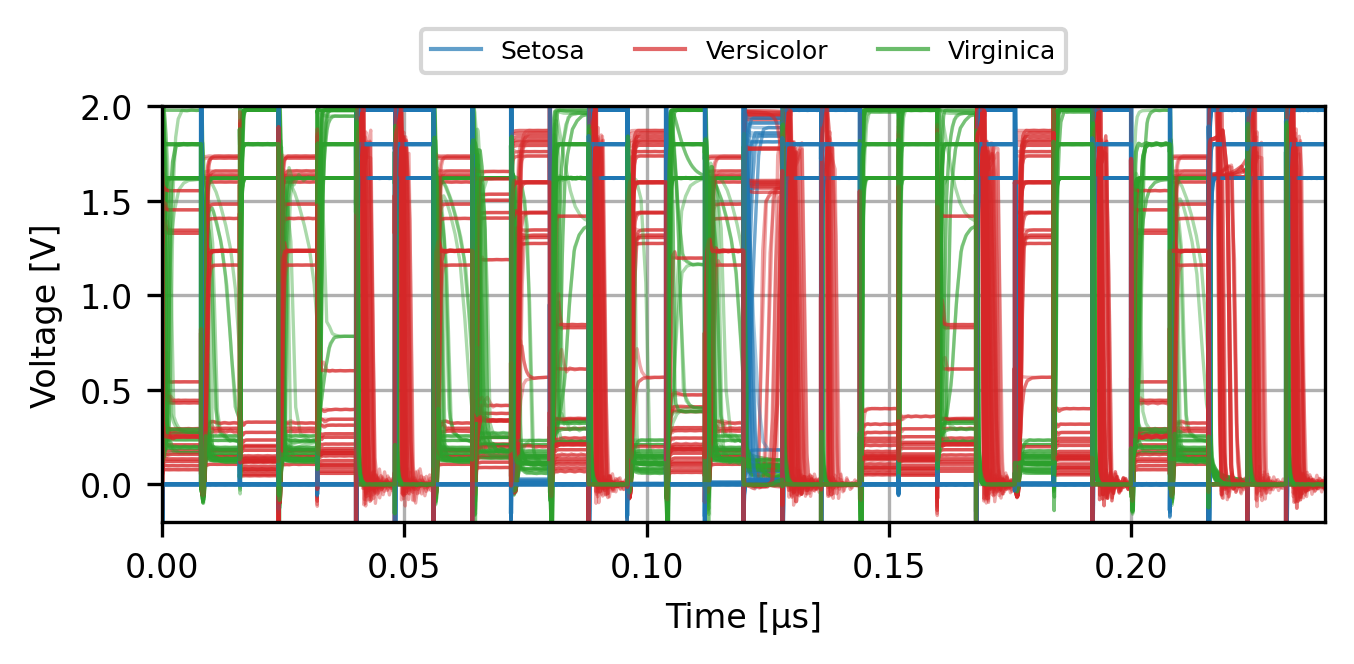}}
        \caption{Output waveforms (best validation circuit)}
    \end{subfigure}
    \begin{subfigure}{\linewidth}
        \centering
        \resizebox{\linewidth}{!}{\includegraphics[trim=0 10 0 0]{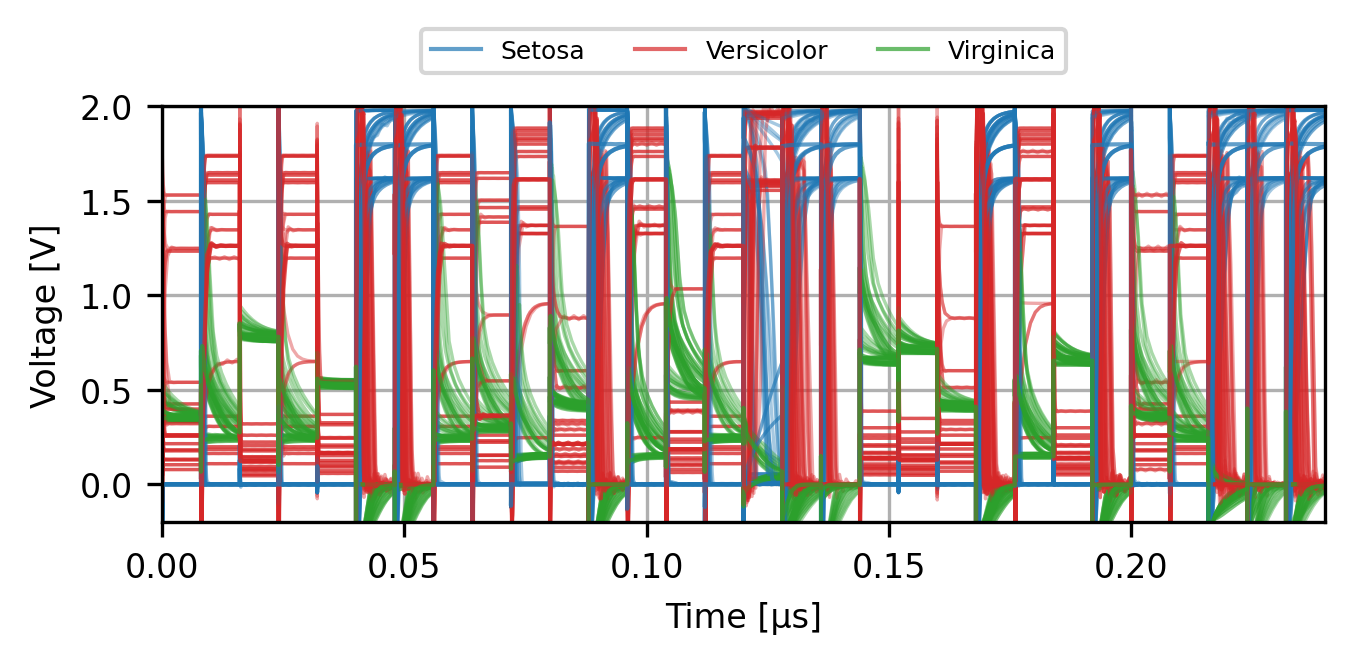}}
        \caption{Output waveforms (second-best validation circuit)}
    \end{subfigure}
    \caption{Voltage transients evaluated on the test split. The plots overlay the waveforms obtained for all corners and all three shuffled versions of the test split.}
    \label{fig:waveforms_corners}
\end{figure}
\end{minipage}

\clearpage

\squeezeSection\section{Robustness to Input-Voltage Perturbations}
\label{appx:test_accuracies}
\begin{figure}[h]
    \centering
    \begin{subfigure}[t]{\linewidth}
        \centering
        \includegraphics[clip,trim=0 55 0 30,width=\linewidth]{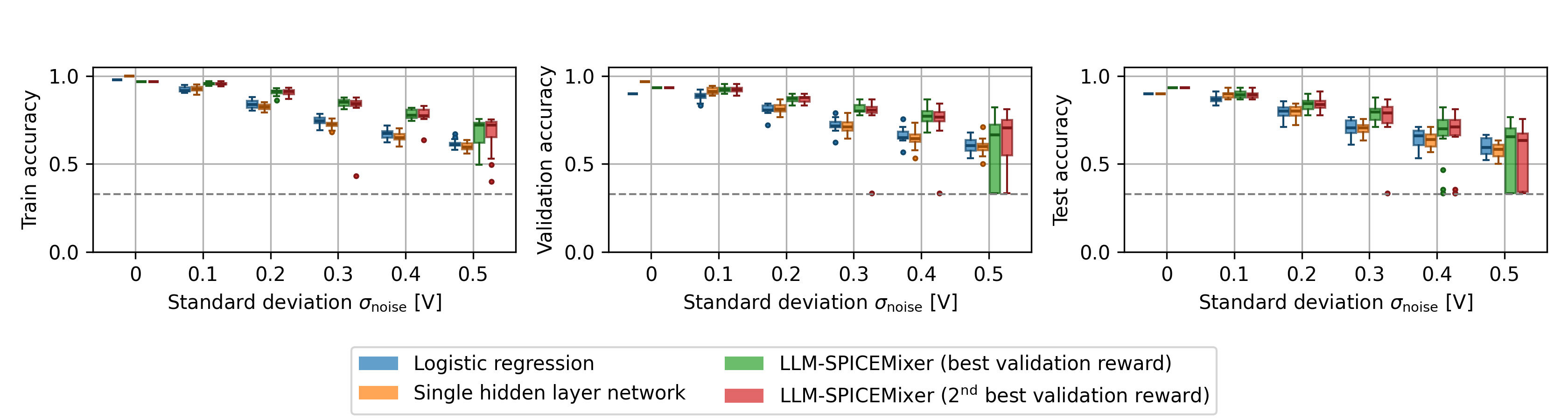}
        \caption{Evaluation over the \texttt{tt} corner}
    \end{subfigure}
    
    \vspace{0.4cm}
    
    \begin{subfigure}[t]{\linewidth}
        \centering
        \includegraphics[clip,trim=0 0 0 30,width=\linewidth]{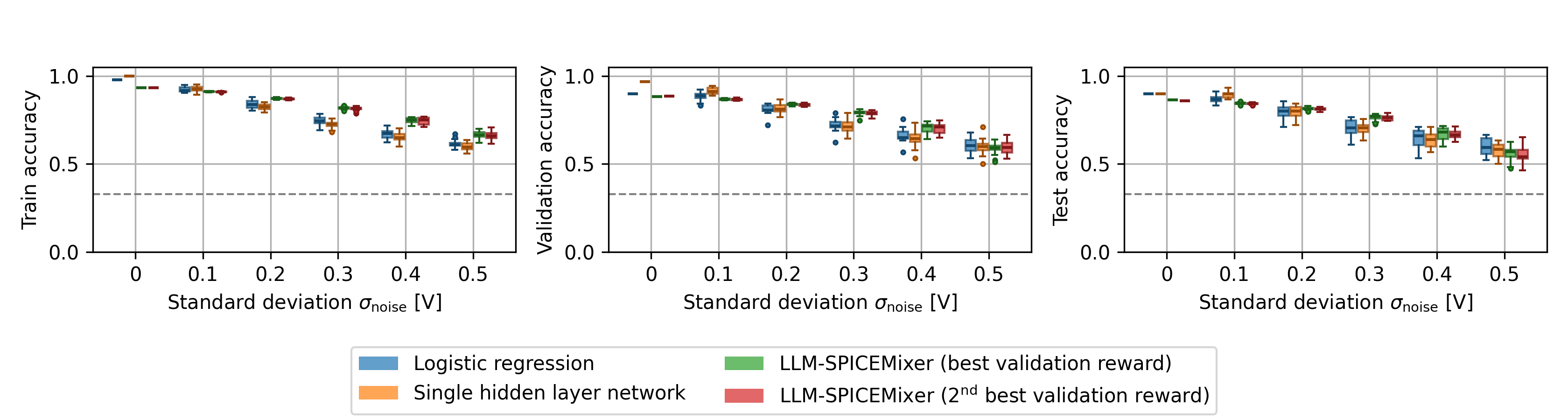}
        \caption{Evaluation over all corners}
    \end{subfigure}
    \vspace{-0.1cm}
    \caption{Robustness of \llmspm{} compared with ML baselines under input-voltage perturbations. The synthesized circuits are competitive with the baselines in the noiseless setting and degrade similarly as Gaussian noise $\sigma_\text{noise}$ is added to the inputs, indicating that the analog solutions retain useful classification margins under imperfect sensor readings. Dashed lines represent chance accuracy.}
    \label{fig:comparison_classifiers}
\end{figure}

Fig.~\ref{fig:comparison_classifiers} and Tab.~\ref{tab:test_noise_results} provide the detailed robustness results for the ML baselines and for the two best validation circuits found by \llmspm{}. We evaluate the classifiers both without input noise and under Gaussian perturbations of the input voltages, with $\sigma_\text{noise} \in \{0, 0.1, 0.2, 0.3, 0.4, 0.5\}$. For each noisy setting, results are aggregated over 16 independent noise realizations.

The boxplots in Fig.~\ref{fig:comparison_classifiers} show that the synthesized circuits are competitive with the ML baselines in the noiseless setting and exhibit a similar degradation trend as the input perturbation increases. Tab.~\ref{tab:test_noise_results} reports the corresponding numerical test accuracies, including mean, standard deviation, and median values. For \llmspm{}, we report both the nominal \texttt{tt}-corner performance and the average performance across all process, voltage, and temperature corners.

\vspace{0.4cm}

\begin{table}[H]
\centering
\caption{Test-set accuracy across noise levels. For \llmspm{}, both \texttt{tt}-only and all-corner results are shown. Each cell reports mean $\pm$ standard deviation / median, in percent ($\%$).}
\label{tab:test_noise_results}
\vspace{-0.3cm}
\resizebox{\linewidth}{!}{\begin{tabular}{llcccccc}
\toprule
Method & Corners & $\sigma_\text{noise}=0$ & $\sigma_\text{noise}=0.1$ & $\sigma_\text{noise}=0.2$ & $\sigma_\text{noise}=0.3$ & $\sigma_\text{noise}=0.4$ & $\sigma_\text{noise}=0.5$ \\
\midrule
Logistic regression
& N/A
& $90.0 \pm 0.0$ / $90.0$
& $87.0 \pm 2.1$ / $86.7$
& $79.5 \pm 3.7$ / $80.0$
& $70.8 \pm 4.6$ / $70.6$
& $64.3 \pm 5.2$ / $66.1$
& $59.8 \pm 5.1$ / $59.4$ \\

Single-hidden-layer network
& N/A
& $90.0 \pm 0.0$ / $90.0$
& $89.6 \pm 1.9$ / $90.0$
& $79.7 \pm 3.2$ / $80.0$
& $69.8 \pm 3.5$ / $70.6$
& $63.7 \pm 4.2$ / $63.9$
& $57.8 \pm 4.0$ / $58.3$ \\
\midrule
\multirow{2}{*}{\llmspm{} (best validation)}
& tt
& $93.3 \pm 0.0$ / $93.3$
& $89.2 \pm 1.9$ / $89.4$
& $84.1 \pm 3.7$ / $83.9$
& $76.0 \pm 12.0$ / $78.9$
& $65.6 \pm 15.7$ / $71.1$
& $55.0 \pm 15.9$ / $63.3$ \\
& all
& $85.9 \pm 0.0$ / $85.9$
& $84.2 \pm 1.3$ / $84.2$
& $81.0 \pm 2.6$ / $81.1$
& $74.8 \pm 3.9$ / $74.4$
& $63.4 \pm 8.5$ / $63.4$
& $52.1 \pm 9.6$ / $53.9$ \\
\midrule
\multirow{2}{*}{\llmspm{} ($2^{\mathrm{nd}}$ best validation)}
& tt
& $93.3 \pm 0.0$ / $93.3$
& $90.0 \pm 2.5$ / $90.0$
& $83.5 \pm 4.1$ / $85.0$
& $78.2 \pm 4.5$ / $77.8$
& $72.9 \pm 4.3$ / $73.3$
& $60.3 \pm 13.9$ / $64.4$ \\
& all
& $87.5 \pm 0.0$ / $87.5$
& $86.1 \pm 1.8$ / $86.3$
& $82.0 \pm 2.8$ / $82.4$
& $76.0 \pm 4.5$ / $76.5$
& $67.5 \pm 7.0$ / $69.7$
& $57.0 \pm 9.0$ / $59.4$ \\
\bottomrule
\end{tabular}}
\end{table}

\clearpage

\squeezeSection\section{Results of Ablation Studies}
\label{appx:ablation_results}
Tabs.~\ref{tab:prompt}--\ref{tab:igel_only} provide the detailed results of the ablation studies for prompting strategy, decoding settings, model choice, and the role of the operator mixture. For each setting, the tables report the final best training reward over nine runs in terms of average, standard deviation, minimum, median, and maximum. In each ablation, we varied only one factor while keeping all other settings identical to the default IGEL configuration. For Tab.~\ref{tab:prompt}, Tab.~\ref{tab:decoding}, and Tab.~\ref{tab:igel_only}, we used Gemma3 12B because of GPU resource constraints.

Tab.~\ref{tab:igel_only} compares IGEL-only search with the full operator mixture. To separate proposal budget from LLM-call budget, we report the full operator mixture both at the reduced proposal budget of $18{,}816$ steps and at the full budget of $131{,}072$ steps. The IGEL-only run uses $18{,}816$ proposal steps, corresponding approximately to the number of LLM calls made in one full \llmspm{} run.

\begin{table}[H]
    \caption{Ablation: final best reward for different prompting strategies on the training split. Values are computed over nine independent runs.}
    \label{tab:prompt}
    \vspace{-0.3cm}
    \centering
    \small
    \renewcommand{\arraystretch}{1.15}
    \setlength{\tabcolsep}{4pt}
    \begin{tabular}{rw{c}{1.9cm}w{c}{1.9cm}w{c}{1.9cm}w{c}{1.9cm}w{c}{1.9cm}w{c}{1.9cm}}
    \toprule
    & \multicolumn{3}{c}{\textbf{without reasoning}} & \multicolumn{3}{c}{\textbf{with reasoning}} \\
    \cmidrule(lr){2-4} \cmidrule(lr){5-7}
     & ``raw''-style & ``diff''-style & alternating & ``raw''-style & ``diff''-style & alternating \\
    \midrule
    \textbf{Average} $\pm$ \textbf{Std. Dev.}
      & \cellcolor{cellred!12} 0.731 $\pm$ 0.031
      & \cellcolor{cellred!5}  0.736 $\pm$ 0.020
      & \cellcolor{cellred!9}  0.733 $\pm$ 0.036
      & \cellcolor{cellred!12} 0.731 $\pm$ 0.038
      & \cellcolor{cellgreen!7} 0.745 $\pm$ 0.045
      & \cellcolor{cellgreen!7} 0.745 $\pm$ 0.042 \\
    \textbf{Minimum}
      & \cellcolor{cellred!53}  0.697
      & \cellcolor{cellred!44}  0.704
      & \cellcolor{cellred!72}  0.681
      & \cellcolor{cellred!83}  0.672
      & \cellcolor{cellred!100} 0.658
      & \cellcolor{cellred!77}  0.677 \\
    \textbf{Median}
      & \cellcolor{cellred!33}   0.713
      & \cellcolor{cellgreen!7}  0.745
      & \cellcolor{cellred!14}   0.729
      & \cellcolor{cellgreen!14} 0.750
      & \cellcolor{cellgreen!20} 0.754
      & \cellcolor{cellgreen!33} 0.763 \\
    \textbf{Maximum}
      & \cellcolor{cellgreen!70}  0.788
      & \cellcolor{cellgreen!24}  0.757
      & \cellcolor{cellgreen!57}  0.779
      & \cellcolor{cellgreen!88}  0.800
      & \cellcolor{cellgreen!91}  0.802
      & \cellcolor{cellgreen!100} 0.808 \\
    \bottomrule
    \end{tabular}
\end{table}

\begin{table}[H]
    \caption{Ablation: final best reward for different decoding settings on the training split. Values are computed over nine independent runs.}
    \label{tab:decoding}
    \vspace{-0.3cm}
    \centering
    \small
    \renewcommand{\arraystretch}{1.15}
    \setlength{\tabcolsep}{4pt}
    \begin{tabular}{rw{c}{1.9cm}w{c}{1.9cm}w{c}{1.9cm}w{c}{1.9cm}w{c}{1.9cm}}
    \toprule
    & \textbf{Deterministic} & \textbf{Conservative} & \textbf{Balanced} & \textbf{Entropy} & \textbf{Entropy++} \\
    & {\scriptsize $T = 0$, $p_\text{top} = 1.0$}
    & {\scriptsize $T = 0.3$, $p_\text{top} = 0.9$}
    & {\scriptsize $T = 0.7$, $p_\text{top} = 0.9$}
    & {\scriptsize $T = 1.0$, $p_\text{top} = 0.95$}
    & {\scriptsize $T = 1.3$, $p_\text{top} = 0.98$}\\
    \midrule
    \textbf{Average} $\pm$ \textbf{Std. Dev.}
      & \cellcolor{cellred!5}   0.740 $\pm$ 0.028
      & \cellcolor{cellred!11}  0.734 $\pm$ 0.037
      & \cellcolor{cellgreen!1} 0.745 $\pm$ 0.042
      & \cellcolor{cellred!20}  0.726 $\pm$ 0.040
      & \cellcolor{cellred!1}   0.744 $\pm$ 0.040 \\
    \textbf{Minimum}
      & \cellcolor{cellred!48}  0.699
      & \cellcolor{cellred!100} 0.650
      & \cellcolor{cellred!71}  0.677
      & \cellcolor{cellred!92}  0.658
      & \cellcolor{cellred!69}  0.679 \\
    \textbf{Median}
      & \cellcolor{cellred!4}   0.741
      & \cellcolor{cellgreen!18} 0.756
      & \cellcolor{cellgreen!29} 0.763
      & \cellcolor{cellgreen!6}  0.748
      & \cellcolor{cellgreen!17} 0.755 \\
    \textbf{Maximum}
      & \cellcolor{cellgreen!84}  0.798
      & \cellcolor{cellgreen!43}  0.772
      & \cellcolor{cellgreen!100} 0.808
      & \cellcolor{cellgreen!42}  0.771
      & \cellcolor{cellgreen!84}  0.798 \\
    \bottomrule
    \end{tabular}
\end{table}

\begin{table}[H]
    \caption{Ablation: final best reward for different models on the training split. Values are computed over nine independent runs.}
    \label{tab:reward_models}
    \vspace{-0.3cm}
    \centering
    \small
    \renewcommand{\arraystretch}{1.15}
    \setlength{\tabcolsep}{5pt}
    \begin{tabular}{rcccccccc}
    \toprule
    & \textbf{Gemma3 270M} & \textbf{Gemma3 1B} & \textbf{Gemma3 4B} & \textbf{Gemma3 12B} & \textbf{Gemma3 27B} & & \textbf{Qwen3.5 9B} & \textbf{Qwen3.5 27B} \\
    \midrule
\textbf{Average} $\pm$ \textbf{Std. Dev.}
      & \cellcolor{cellred!12}   0.737 $\pm$ 0.034
      & \cellcolor{cellred!8}    0.740 $\pm$ 0.039
      & \cellcolor{cellgreen!2}  0.746 $\pm$ 0.028
      & \cellcolor{cellgreen!1}  0.745 $\pm$ 0.042
      & \cellcolor{cellgreen!6}  0.750 $\pm$ 0.019
      &
      & \cellcolor{cellred!3}    0.743 $\pm$ 0.040
      & \cellcolor{cellgreen!35} 0.799 $\pm$ 0.040 \\
    \textbf{Minimum}
      & \cellcolor{cellred!58}   0.686
      & \cellcolor{cellred!100}  0.671
      & \cellcolor{cellred!31}   0.701
      & \cellcolor{cellred!83}   0.677
      & \cellcolor{cellred!18}   0.709
      &
      & \cellcolor{cellred!100}  0.671
      & \cellcolor{cellgreen!12} 0.719 \\
    \textbf{Median}
      & \cellcolor{cellred!4}     0.737
      & \cellcolor{cellgreen!11}  0.749
      & \cellcolor{cellgreen!3}   0.744
      & \cellcolor{cellgreen!35}  0.763
      & \cellcolor{cellgreen!30}  0.760
      &
      & \cellcolor{cellgreen!12}  0.750
      & \cellcolor{cellgreen!100} 0.810 \\
    \textbf{Maximum}
      & \cellcolor{cellgreen!63}  0.801
      & \cellcolor{cellgreen!24}  0.790
      & \cellcolor{cellgreen!29}  0.793
      & \cellcolor{cellgreen!71}  0.808
      & \cellcolor{cellgreen!1}   0.777
      &
      & \cellcolor{cellgreen!78}  0.812
      & \cellcolor{cellgreen!100} 0.855 \\
    \bottomrule
    \end{tabular}
\end{table}

\begin{table}[H]
    \caption{Ablation: final best reward for Gemma3 12B when using the full operator mixture (= LLM-SPICEMixer) versus using only IGEL. The first two columns compare both settings at the same number of proposal steps. The last column reports the full all-operator run for reference; the IGEL-only setting uses approximately the same number of LLM calls as this full run.}
    \label{tab:igel_only}
    \vspace{-0.3cm}
    \centering
    \small
    \renewcommand{\arraystretch}{1.15}
    \setlength{\tabcolsep}{5pt}
    \begin{tabular}{rw{c}{2.9cm}w{c}{2.9cm}w{c}{2.9cm}}
    \toprule
    & \makecell{\textbf{All operators}\\{\footnotesize $18{,}816$ steps}\\{\scriptsize ($16{,}128$ SPICEMixer + $2{,}688$ IGEL)}}
    & \makecell{\textbf{Only IGEL}\\{\footnotesize $18{,}816$ steps}\\{\scriptsize ($0$ SPICEMixer + $18{,}816$ IGEL)}}
    & \makecell{\textbf{All operators}\\{\footnotesize $131{,}072$ steps}\\{\scriptsize ($112{,}347$ SPICEMixer + $18{,}725$ IGEL)}} \\
    \midrule
    \textbf{Average} $\pm$ \textbf{Std. Dev.}
      & \cellcolor{cellgreen!24}  0.705 $\pm$ 0.031
      & \cellcolor{cellred!49}    0.607 $\pm$ 0.035
      & \cellcolor{cellgreen!53}  0.745 $\pm$ 0.042 \\
    \textbf{Minimum}
      & \cellcolor{cellred!6}     0.665
      & \cellcolor{cellred!100}   0.538
      & \cellcolor{cellgreen!3}   0.677 \\
    \textbf{Median}
      & \cellcolor{cellgreen!12}  0.689
      & \cellcolor{cellred!40}    0.619
      & \cellcolor{cellgreen!67}  0.763 \\
    \textbf{Maximum}
      & \cellcolor{cellgreen!53}  0.744
      & \cellcolor{cellred!13}    0.656
      & \cellcolor{cellgreen!100} 0.808 \\
    \bottomrule
    \end{tabular}
\end{table}

\squeezeSection\section{LLM Response Length Statistics}
\label{appx:response_length_statistics}
Tab.~\ref{tab:response_length_stats} summarizes the response lengths of the evaluated LLMs in terms of both characters and words. For the Qwen3.5 models, we report thinking tokens and final output separately. These results provide additional context for the model comparison discussed in the main paper.

\vspace{1cm}

\begin{table}[H]
    \caption{Response-length statistics across the four models. For the Qwen3.5 models, values are reported as \texttt{<think>} $+$ final output, because these models produce explicit thinking tokens.}
    \label{tab:response_length_stats}
    \vspace{-0.3cm}
    \centering
    \small
    \renewcommand{\arraystretch}{1.15}
    \setlength{\tabcolsep}{6pt}
    \begin{tabular}{rw{c}{2.9cm}w{c}{2.9cm}w{c}{2.9cm}w{c}{2.9cm}}
    \toprule
    & \textbf{Gemma3 12B} & \textbf{Gemma3 27B} & \textbf{Qwen3.5 9B} & \textbf{Qwen3.5 27B} \\
    \midrule
    \multicolumn{5}{c}{\emph{Number of characters}} \\
    \textbf{Average} $\pm$ \textbf{Std. Dev.}
        & $1{,}576.7 \pm 389.2$
        & $1{,}215.9 \pm 336.0$
        & \makecell{$5{,}401.5 \pm 4{,}217.8\quad$\\ $\quad+~1{,}029.4 \pm 625.6$}
        & \makecell{$21{,}825.9 \pm 8{,}690.7\quad$\\ $\quad+~1{,}322.9 \pm 725.3$} \\
    \textbf{Minimum}
        & 430
        & 263
        & $1{,}173 + 61$
        & $1{,}216 + 62$ \\
    \textbf{Median}
        & 1,532
        & 1,176
        & $3{,}961 + 894$
        & $23{,}871 + 1{,}336$ \\
    \textbf{Maximum}
        & 3,780
        & 16,977
        & $59{,}653 + 23{,}750$
        & $69{,}535 + 23{,}088$ \\
    \midrule
    \multicolumn{5}{c}{\emph{Number of words}} \\
    \textbf{Average} $\pm$ \textbf{Std. Dev.}
        & $226.2 \pm 54.1$
        & $177.1 \pm 46.8$
        & \makecell{$793.1 \pm 627.2\quad$\\ $\quad+~141.1 \pm 89.8$}
        & \makecell{$3{,}368.5 \pm 1{,}364.4\quad$\\ $\quad+~186.4 \pm 102.9$} \\
    \textbf{Minimum}
        & 65
        & 42
        & $137 + 10$
        & $170 + 10$ \\
    \textbf{Median}
        & 221
        & 172
        & $576 + 121$
        & $3{,}675 + 189$ \\
    \textbf{Maximum}
        & 498
        & 2,333
        & $9{,}121 + 3{,}094$
        & $7{,}205 + 3{,}170$ \\
    \bottomrule
    \end{tabular}
\end{table}

\end{document}